%% file: main.tex
\documentclass[sigconf]{acmart}

\AtBeginDocument{%
  }

\copyrightyear{2025}
\acmYear{2025}
\setcopyright{cc}
\setcctype{by}
\acmConference[KDD '25]{Proceedings of the 31st ACM SIGKDD Conference on
Knowledge Discovery and Data Mining V.2}{August 3--7, 2025}{Toronto, ON,
Canada}
\acmBooktitle{Proceedings of the 31st ACM SIGKDD Conference on Knowledge
Discovery and Data Mining V.2 (KDD '25), August 3--7, 2025, Toronto, ON,
Canada}
\acmDOI{10.1145/3711896.3737122}
\acmISBN{979-8-4007-1454-2/2025/08}

\input{dfn.tex}
\begin{document}

\title{Self-Tuning Self-Supervised Image Anomaly Detection}

\author{Jaemin Yoo}
\affiliation{%
  \institution{KAIST}
  \city{Daejeon}
  \country{South Korea}
}
\email{jaemin@kaist.ac.kr}

\author{Lingxiao Zhao}
\affiliation{%
  \institution{Mistral AI}
  \city{Palo Alto}
  \state{CA}
  \country{USA}
}
\email{lingxiaozlx@gmail.com}

\author{Leman Akoglu}
\affiliation{%
  \institution{Carnegie Mellon University}
  \city{Pittsburgh}
  \state{PA}
  \country{USA}
}
\email{lakoglu@andrew.cmu.edu}

\renewcommand{\shortauthors}{Jaemin Yoo, Lingxiao Zhao, and Leman Akoglu}

\begin{abstract}
Self-supervised learning (SSL) has emerged as a promising paradigm that presents supervisory signals to real-world problems, bypassing the extensive cost of manual labeling.
Consequently, self-supervised anomaly detection (SSAD) has seen a recent surge of interest, since SSL is especially attractive for unsupervised tasks.
However, recent works have reported that the choice of a data augmentation function has significant impact on the accuracy of SSAD, posing \textit{augmentation search} as an essential but nontrivial problem due to lack of labeled validation data.
In this paper, we introduce \method, the \textit{first unsupervised approach to end-to-end augmentation tuning for SSAD}.
To this end, our work presents two key contributions.
The first is a new unsupervised validation loss that quantifies the alignment between augmented training data and unlabeled validation data.
The second is new differentiable augmentation functions, allowing data augmentation hyperparameter(s) to be tuned in an end-to-end manner.
Experiments on two testbeds with semantic class anomalies and subtle industrial defects show that \method gives significant performance gains over existing works.
All our code and testbeds are available at \url{https://github.com/jaeminyoo/ST-SSAD}.
\end{abstract}





\maketitle

\input{010intro}
\input{020problem}
\input{030related}
\input{040method}

\input{050experiments}
\input{060conclusion}


\begin{acks}
This work was partly supported by the National Research Foundation of Korea (NRF) grant funded by the Korea government (MSIT) (RS-2024-00341425) and NSF IIS 2310482.
\end{acks}

\bibliographystyle{ACM-Reference-Format}
\bibliography{main}


\appendix
\input{100appendix}

%
%
%
%
%
%
%
%

\end{document}

%% file: dfn.tex
\usepackage{graphicx}
\usepackage{xspace}
\usepackage{subcaption}
\usepackage{bm}
\usepackage{amsmath}
\usepackage{wrapfig}
\usepackage{paralist}
\usepackage{multirow}
\usepackage{colortbl}
\usepackage{algorithm}
\usepackage{algorithmic}
\usepackage{xcolor}

\newcommand{\cellFirst}[1]{\cellcolor{green}{#1}}
\newcommand{\cellSecond}[1]{\cellcolor{green!50}{#1}}
\newcommand{\cellThird}[1]{\cellcolor{green!25}{#1}}

\newcommand{\trnData}[0]{\mathcal{D}_\mathrm{trn}}
\newcommand{\augData}[0]{\mathcal{D}_\mathrm{aug}}
\newcommand{\testData}[0]{\mathcal{D}_\mathrm{val}}
\newcommand{\testDataN}[0]{\mathcal{D}_\mathrm{val}^{{\scriptscriptstyle (}\mathrm{n}{\scriptscriptstyle )}}}
\newcommand{\testDataA}[0]{\mathcal{D}_\mathrm{val}^{{\scriptscriptstyle (}\mathrm{a}{\scriptscriptstyle )}}}

\newcommand{\trnEmb}[0]{\mathcal{Z}_\mathrm{trn}}
\newcommand{\augEmb}[0]{\mathcal{Z}_\mathrm{aug}}
\newcommand{\testEmb}[0]{\mathcal{Z}_\mathrm{val}}

\newcommand{\testEmbN}[0]{\mathcal{Z}_\mathrm{val}^{{\scriptscriptstyle (}\mathrm{n}{\scriptscriptstyle )}}}
\newcommand{\testEmbA}[0]{\mathcal{Z}_\mathrm{val}^{{\scriptscriptstyle (}\mathrm{a}{\scriptscriptstyle )}}}

\newcommand{\trnLoss}[0]{\mathcal{L}_\mathrm{trn}}
\newcommand{\valLoss}[0]{\mathcal{L}_\mathrm{val}}
\newcommand{\baseLoss}[0]{\smash{\mathcal{L}_\mathrm{val}^{(\mathrm{b})}}}

\newcommand{\detector}[0]{f_\theta}
\newcommand{\augFunc}[0]{\phi_\mathrm{aug}}
\newcommand{\augParam}[0]{\mathbf{a}}

\newcommand{\myParagraph}[1]{\textbf{#1.} }

\newcommand{\method}{ST-SSAD\xspace}
\newcommand{\numData}{41\xspace}

%% file: 010intro.tex
\section{Introduction}
\label{sec:intro}

Anomaly detection (AD) finds various applications in security, finance, and manufacturing, to name a few.
Thanks to its popularity, the literature is abound with numerous detection techniques \citep{bookCharu}, while neural network-based models have attracted the most attention recently \citep{pang2021deep}.
Especially for dynamically-changing settings in which anomalies are to be identified, it is important to design \textit{unsupervised} techniques.
While supervised detection can be employed for label-rich settings, unsupervised detection becomes critical to remain alert to the so-called ``unknown unknowns''.

Recently, self-supervised learning (SSL) has emerged as a promising paradigm that offers supervisory
signals to real-world problems while avoiding the extensive cost of manual labeling, leading to great success in advancing NLP \citep{conneau2020unsupervised, brown2020language} as well as computer vision tasks \citep{goyal2021self, he2022masked}.
SSL has become particularly attractive for \textit{unsupervised} problems such as AD, where labeled data is either nonexistent, costly to obtain, or nontrivial to simulate in the face of unknown anomalies.
For this reason, the literature has seen a recent surge of SSL-based AD (SSAD) techniques \citep{Golan18GEOM,Hendrycks19OE,Bergman20GOAD,Li21CutPaste,Sehwag21SSD,Qiu21NeuTraL}.
The typical approach to SSAD involves incorporating {self-generated} \textit{pseudo}
anomalies into training, and then learning to separate those from the inliers.
The pseudo anomalies are most often synthesized artificially by transforming inliers through a data augmentation function, such as masking, blurring, or rotation.

In this paper, we address a fundamental challenge in SSAD, to which recent works seem to have turned a blind eye: recognizing and tuning \textit{augmentation as a hyperparameter}.
As shown recently by \citet{yoo2022role} through extensive experiments, the choice of a data augmentation function, as well as its associated arguments such as the masking amount, blurring level, etc., has a tremendous impact on detection accuracy.
This may come across as no surprise since the supervised learning community already has integrated augmentation hyperparameters into model selection \citep{Cubuk19Auto,ottoni2023tuning}.
Meanwhile, there exist few attempts in the literature on SSAD.
Although model selection without labels is admittedly a much harder problem, turning a blind eye to the challenge may mislead by overstating the (unreasonable) effectiveness of SSL for unsupervised AD.

\begin{figure*}
    \centering
    \includegraphics[width=0.84\textwidth, trim=10mm 0 11mm 0, clip]{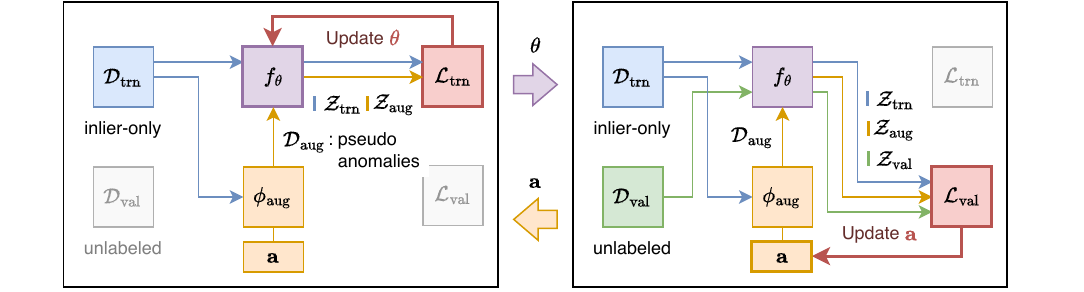}
    \vspace{1mm}
    \caption{
        (left) Training  and (right) validation (i.e., augmentation tuning) stages of \method, which alternates between: (left) Given augmentation function $\augFunc(\cdot ; \mathbf{a})$, estimate the parameters $\theta$ of detector $\detector$ on inliers $\trnData$ and pseudo anomalies $\augData$ via the training loss;
        (right) Given $f_\theta$ and unlabeled validation data $\testData$, update the augmentation hyperparameters $\augParam$ through our unsupervised validation loss that measures the agreement between $\trnData \cup \augData$ and $\testData$ in the embedding space.
    }
    \label{fig:overview}
\end{figure*}

Recent work \cite{YooZZA23} addressed the augmentation search problem on SSAD by designing an unsupervised validation loss that can compare models with different configurations without labels.
The main limitation is that all models with different configurations need to be trained before the validation loss is applied, which causes extensive computational time exponential with the number of hyperparameters to tune.
Thus, it cannot save time required for hyperparameter search, and its practical usability is severely limited.

Our work introduces \method, the \emph{first systematic approach for rigorous augmentation tuning on SSAD}.
Intuitively, SSAD works well if the generated pseudo anomalies resemble the true anomalies.
Put differently, this is when the augmentation function well mimics the (unknown) anomaly-generating mechanism.
Based on this insight, \textbf{(1)} we design a new unsupervised validation loss for SSAD toward quantifying the alignment between the augmented training data and \emph{unlabeled} validation data. 
The loss is differentiable, allowing us to tune augmentation in an end-to-end manner (see Fig.~\ref{fig:overview}).
This necessitates the augmentation function to be differentiable as well.
To this end, \textbf{(2)} we propose new differentiable formulations for popular augmentations such as CutOut \citep{Devries17CutOut} and rotation \citep{Golan18GEOM} as a proof of concept, which aim at gross (semantic) and local (subtle) anomalies, respectively, following previous works \citep{Golan18GEOM, Li21CutPaste}.

Our experiments on 41 anomaly detection tasks including both local and global anomalies demonstrate that \method significantly outperforms both unsupervised and self-supervised baselines which rely on manual hyperparameter search without labels.
Our qualitative analysis visually supports that \method is capable of learning appropriate augmentation hyperparameters for different anomaly types, even when they share the same normal data, by leveraging the anomalies in unlabeled validation data.
We also demonstrate the linear complexity of \method both theoretically and empirically, making sure that it can be directly applicable to large data.

%% file: 020problem.tex
\section{Notation and Problem Definition}
\label{sec:prelim}

\myParagraph{Notation}
Let $\trnData$ represent a set of training normal data (i.e. inliers), and $\testData$ be a set of unlabeled validation data containing both normal and anomalous samples.
Let $\mathbf{x} \in \mathbb{R}^d$ denote a data sample with size $d$.
Let $\augFunc \in \mathbb{R}^d \times \mathcal{A} \mapsto \mathbb{R}^d$ depict a data augmentation function conditioned on its hyperparameters $\mathbf{a} \in \mathcal{A}$, where $\mathcal{A}$ is the set of all possible values.
For example, if $\augFunc$ is the rotation of an image, $\mathcal{A} = [0, 360)$ is the set of angles, and $\augFunc (\mathbf{x}; a)$ is the image rotated by angle $a$.
Let \smash{$\detector \in \mathbb{R}^d \mapsto \mathbb{R}^h$} be a detector parameterized by $\theta$, and \smash{$s \in \mathbb{R}^h \mapsto \mathbb{R}^+$} be an anomaly score function.
During test time, we decide $\mathbf{x}$ as an anomaly if $s(f_\theta(\mathbf{x}))$ is high.
Specifically, $f_\theta$ returns a low-dimensional embedding $\mathbf{z} \in \mathbb{R}^h$ for $\mathbf{x}$, which is then fed into $s$ to compute the score of $\mathbf{x}$.
We assume that $f_\theta$ is trained in a self-supervised way by using a set $\augData = \{ \augFunc(\mathbf{x}; \augParam) \mid \mathbf{x} \in \trnData \}$ of pseudo anomalies as done in previous works \cite{Li21CutPaste}.

\myParagraph{Problem definition}
Given $\trnData$ and $\testData$, how can we efficiently find  $\augParam^*$ (along with the model parameters $\theta$) that maximizes the accuracy of the detector $f_\theta$ with score function $s$?

There is no trivial solution to the problem, mainly because labeled anomalies are non-existent at training time---as labeled data is often hard to gather in anomaly detection. 
However, the problem is crucial for SSAD in real-world tasks since the choice of such hyperparameters not only affects the performance of anomaly detection, but determines the success of it. 
To our knowledge, there is no end-to-end approach for solving the problem in the literature, and our work is the first to propose a systematic solution.

%% file: 030related.tex
\section{Related Work}
\label{sec:related}

\myParagraph{Self-supervised learning (SSL)}
SSL has seen a surge of attention, especially for foundation models \citep{bommasani2021opportunities} like LLMs, which can generate remarkable human-like text \citep{zhou2023comprehensive}.
Self-supervised representation learning has also offered an astonishing boost to a variety of tasks in NLP, vision, and recommendation \citep{liu2021self}.
SSL has been argued as the key toward ``unlocking the dark matter of intelligence'' \citep{darkmatter21}.

\myParagraph{Self-supervised anomaly detection (SSAD)}
In terms of the pretext task and the associated loss on which they are trained, most SSL methods can be categorized as generative or contrastive.
Generative SSAD can further be organized based on (denoising) autoencoders \citep{Zhou17RDA,Zong18DAGMM, Cheng21TIAE, Ye22ARNet}, adversarial models \citep{Akcay18GANomaly,Zenati18ALAD}, and flow-based models \citep{Rudolph22CSFlow,Gudovskiy22CFlowAD}.
On the other hand, contrastive SSAD relies on data augmentation that generates pseudo anomalies by transforming inliers, and a supervised loss for distinguishing between inliers and the pseudo anomalies \citep{hojjati2022self}.
Data augmentation functions for contrastive SSAD include geometric \citep{Golan18GEOM, Bergman20GOAD}, CutPaste \citep{Li21CutPaste}, patch-wise cloning \cite{Schlter2021SelfSupervisedOD}, distribution-shifting transformation \citep{Tack20CSI}, masking \citep{ijcai2021p198}, and learnable neural network-based transformation \citep{Qiu21NeuTraL}.

Data augmentation functions can be further categorized based on the nature of target anomalies: subtle (local) and semantic (global) anomalies.
Subtle anomalies appear in a part of an image and are well-modeled by local augmentation functions such as CutOut~\citep{Devries17CutOut} or CutPaste~\citep{Li21CutPaste}.
Semantic anomalies are better characterized with the semantic difference rather than pixel-level differences, and are well-modeled by global transformation functions.
Specifically, previous works have shown that geometric functions \citep{Golan18GEOM, Bergman20GOAD} such as rotation are effective for capturing semantic anomalies.
In this work, we devise differentiable variants of CutOut and rotation functions, respectively, to deal with both types of anomalies.

\myParagraph{Automating augmentation}
Recent work in computer vision (CV) has shown that the success of SSL relies on proper augmentation functions \citep{steiner2022how,touvron2022deit}.
Sensitivity to the choice of augmentation has also been shown for SSAD recently \citep{yoo2022role}. 
While data augmentation in CV aims at improving generalization by accounting for invariances, augmentation in SSAD plays the key role of presenting the classifier with specific kinds of pseudo anomalies. 
While the supervised CV community proposed methods toward automating augmentation \citep{Cubuk19Auto,cubuk2020randaugment}, our work is the first attempt toward rigorously tuning data augmentation for SSAD in an end-to-end way.
The key difference is that the former sets aside a \textit{labeled validation} set, whereas we address the arguably more challenging setting for \textit{fully unsupervised} anomaly detection without any labels.

\myParagraph{Data quality metrics}
Recent works \citep{jiang2023opendataval, perini2024uncertainty} proposed metrics for measuring the quality of samples.
One challenge in using these methods for unsupervised model selection is the several valuation options they offer, the selection of which is not obvious.
Moreover, EAP \citep{perini2024uncertainty} requires a set of auxiliary samples as input for conducting model selection, making it challenging to construct this set to choose from upfront.
On the contrary, our work focuses on directly estimating realistic augmentation that best mimics the observed ones \emph{on-the-fly}, without the need for any auxiliary dataset.

%% file: 040method.tex
\section{Proposed Approach}
\label{sec:framework}

We propose \method, a framework for augmentation tuning in self-supervised anomaly detection (SSAD).
Given unlabeled validation data $\testData$, \method automatically finds the best augmentation hyperparameter $\augParam^*$ that maximizes the semantic alignment between the data augmentation and underlying anomaly-generating mechanism in $\testData$.
The search process is performed in an end-to-end fashion thanks to two core novel engines of \method: (1) an \textit{unsupervised validation} loss $\valLoss$ and (2) a \textit{differentiable augmentation} function $\augFunc$, which we describe in Sec. \ref{ssec:valloss} and \ref{ssec:diffaug}, resp.

\begin{algorithm}[t]
    \caption{\method for Augmentation Tuning}
    \begin{algorithmic}[1]
        \REQUIRE Training data $\trnData$,
        unlabeled validation data $\testData$,
        aug. function $\augFunc$,
        training loss $\trnLoss$,
        validation loss $\valLoss$,
        detector $f_\theta$,
        number of epochs $T$, 
        and
        step sizes $\alpha$ and $\beta$
        \ENSURE Optimized augmentation hyperparameters $\augParam^{(T)}$
        \STATE $\augParam^{(0)} \leftarrow $ Initialize augmentation hyperparameters
        \FOR{$t \in \{ 0, 1, \cdots, T-1 \}$}
            \STATE Let \smash{$\theta^{(t+1)} = \theta^{(t)} - \alpha \nabla_{\theta^{(t)}} \mathcal{L}_\mathrm{trn}(\theta^{(t)}, \augParam^{(t)})$}
            \STATE $\trnEmb \leftarrow \{ f_{\theta^{(t+1)}}(\mathbf{x}) \mid \mathbf{x} \in \trnData \}$
            \STATE $\testEmb \leftarrow \{ f_{\theta^{(t+1)}}(\mathbf{x}) \mid \mathbf{x} \in \testData \}$
            \STATE \smash{$\augEmb \leftarrow \{ f_{\theta^{(t+1)}}(\augFunc(\mathbf{x}; \augParam^{(t)})) \mid \mathbf{x} \in \trnData \}$}
            \STATE $\trnEmb, \augEmb, \testEmb \leftarrow \mathrm{normalize}(\trnEmb, \augEmb, \testEmb)$
            \STATE \smash{$\augParam^{(t+1)} \leftarrow \augParam^{(t)} - \beta \nabla_{\augParam^{(t)}} \mathcal{L}_\mathrm{val}(\trnEmb, \augEmb, \testEmb)$}
        \ENDFOR
    \end{algorithmic}
    \label{alg:framework}
\end{algorithm}

Fig. \ref{fig:overview} shows an overall structure of \method, which updates the parameters $\theta$ of the detector $\detector$ and the augmentation hyperparameters $\augParam$ through alternating stages for training and validation.
Let $\trnEmb = \{\detector(\mathbf{x}) \mid \mathbf{x} \in \trnData \}$ be the low-dimensional embeddings of $\trnData$, $\augEmb = \{ \detector(\augFunc(\mathbf{x}; \augParam)) \mid \mathbf{x} \in \trnData \}$ be the embeddings of augmented data, and $\testEmb = \{\detector(\mathbf{x}) \mid \mathbf{x} \in \testData \}$ be the embeddings of $\testData$.
In the training stage, \method updates $\theta$ to minimize the training loss based on the self-supervised task determined by the current $\augFunc$ and $\augParam$.
In the validation stage, \method updates $\augParam$ to reduce the unsupervised validation loss based on the new embeddings generated by the updated $\detector$.
The framework halts when $\mathbf{a}$ reaches a local optimum, typically after a few iterations.

The detailed process of \method is shown as Algorithms \ref{alg:framework}.
Line 3 denotes the training stage, and Lines 4 to 8 represent the validation stage.
The computational chain between $\theta^{(t)}$ and $\theta^{(t+1)}$ remains until Line 8 to support the second-order optimization (Sec. \ref{ssec:implement}).
We execute Algorithm \ref{alg:framework} for multiple random initializations of $\augParam$ (Sec. \ref{ssec:implement}) due to its gradient-based nature; we search for local optima to the bi-level optimization problem given each initialization.

\input{041valloss}
\input{042diffaug}
\input{043impl}
\input{044complexity}

%% file: 041valloss.tex
\subsection{Unsupervised Validation Loss}
\label{ssec:valloss}

The validation loss $\valLoss$ is the first core block of \method, whose goal is to quantify the agreement between $\augFunc$ and the underlying anomaly-generating mechanism even without any labels.
Our idea is to measure the set distance between $\trnData \cup \augData$ and $\testData$, based on the intuition that the two sets will become similar if $\augData$ more resembles the unlabeled anomalies in $\testData$.
Specifically, we use the low-dimensional embeddings instead of the raw data to focus on their semantic representation, avoiding the high dimensionality.

Based on the idea, we present the basic form of our validation loss as $\baseLoss(\trnEmb, \augEmb, \testEmb) = \mathrm{dist}(\trnEmb \cup \augEmb, \testEmb)$, where $\mathrm{dist}(\cdot,\cdot)$ is any function that measures a distance between two sets of vectors.
The three sets $\trnEmb$, $\augEmb$, and $\testEmb$ of embeddings are created from $\trnData$, $\augData$, and $\testData$, respectively, by $\detector$.

\begin{figure}[t]
    \centering
    \includegraphics[width=0.33\textwidth, trim=15mm 9mm 23mm 5mm, clip]{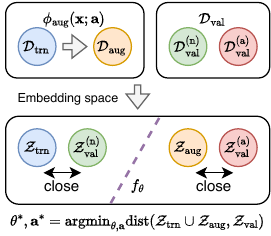}
    \caption{
        Main idea of our proposed validation loss $\valLoss$.
        The loss works as a proxy of the distance between $\augEmb$ and \smash{$\testEmbA$}, which cannot be computed directly without labels.
        By learning the parameters $\theta$ and $\augParam$ that minimize the loss, we align the decision boundary with the validation anomalies.
    }
    \label{fig:val-loss-example}
\end{figure}

Fig. \ref{fig:val-loss-example} illustrates our intuition for designing the validation loss.
Let \smash{$\testDataN$} and \smash{$\testDataA$} be the inliers and anomalies in $\testData$, respectively.
We can consider that the basic loss \smash{$\baseLoss$} measures the two distances between $\trnEmb$ and \smash{$\testEmbN$} and between $\augEmb$ and \smash{$\testEmbA$} and combines them into a scalar.
If we focus on the inliers, we can safely assume that the distance between $\trnEmb$ and \smash{$\testEmbN$} is sufficiently small, since $\trnData$ and \smash{$\testDataN$} are drawn from the same data distribution.
Then, the loss becomes a proxy of the remaining distance between $\augEmb$ and \smash{$\testEmbA$}, which cannot be measured without labels.
As a result, by minimizing \smash{$\baseLoss$}, we can find an augmentation function $\augFunc$ that allows us to use the decision boundary between $\trnEmb$ and $\augEmb$ to separate the unseen inliers and anomalies in the test data.

The choice of $\mathrm{dist}(\cdot,\cdot)$ determines the effectiveness of $\baseLoss$.
We carefully design it to address two notable challenges:
\begin{compactenum}[\textbullet]
    \item \textbf{Scale invariance:}
        During optimization, the scale of embedding distances can arbitrarily change as the augmentation hyperparameters are updated, since the detector model changes accordingly.
        Thus, $\valLoss$ should be robust to the scale of distances if the (relative) distribution of embeddings is preserved.
    \item \textbf{Ratio robustness:}
        Let $\gamma = |\augData| / |\trnData|$ represent the relative size of augmented data, which is basically the number of times we apply $\augFunc$ to $\trnData$ for augmentation.
        Since the anomaly ratio in $\testData$ is unknown, $\valLoss$ should be robust to the value of $\gamma$ which we manually set prior to training.
\end{compactenum}

\myParagraph{Total distance normalization}
For scale invariance, we propose total distance normalization to control the total pairwise squared distance (TPSD) \citep{Zhao20PairNorm} of embeddings.
Let $\mathbf{Z}$ be the embedding matrix that stacks all embeddings in $\trnEmb$, $\augEmb$, and $\testEmb$ as its rows.
Then, the TPSD is defined as \smash{$\mathrm{TPSD}(\mathbf{Z}) = \sum_{ij} \| \mathbf{z}_i - \mathbf{z}_j \|_2^2$}, where $\mathbf{z}_i$ is the $i$-th row of $\mathbf{Z}$.
We transform $\mathbf{Z}$ to have the unit TPSD in linear time as suggested in \citep{Zhao20PairNorm}:
$\mathbf{z}_i' = \frac{\sqrt{N}}{\| \mathbf{Z}^c \|_\mathrm{F}} \mathbf{z}_i^c,$ where \smash{$\mathbf{z}_i^c = \mathbf{z}_i - \frac{1}{N} \sum_{j=1}^N \mathbf{z}_j$,}
$N$ is the number of rows in $\mathbf{Z}$, and $\| \cdot \|_\mathrm{F}$ refers to the Frobenius norm of a matrix.
We stack $\mathbf{z}_i'$ as rows to create $\mathbf{Z}'$.

By using $\mathbf{Z}'$ instead of $\mathbf{Z}$ for computing the distances, we focus on the relative distances between embeddings while maintaining the overall variance.
It is noteworthy that the vector normalization, i.e., $\mathbf{z}_i \leftarrow \mathbf{z}_i / \| \mathbf{z}_i \|_2 \ \forall i$, does not solve the scale invariance problem since the scale of distances can still be arbitrary even on unit vectors.
Another advantage of total distance normalization is that it steers away from the trivial solution of the distance minimization problem, which is to set all embeddings to the zero vector.

\begin{figure}[t]
    \centering
    \includegraphics[width=0.32\textwidth]{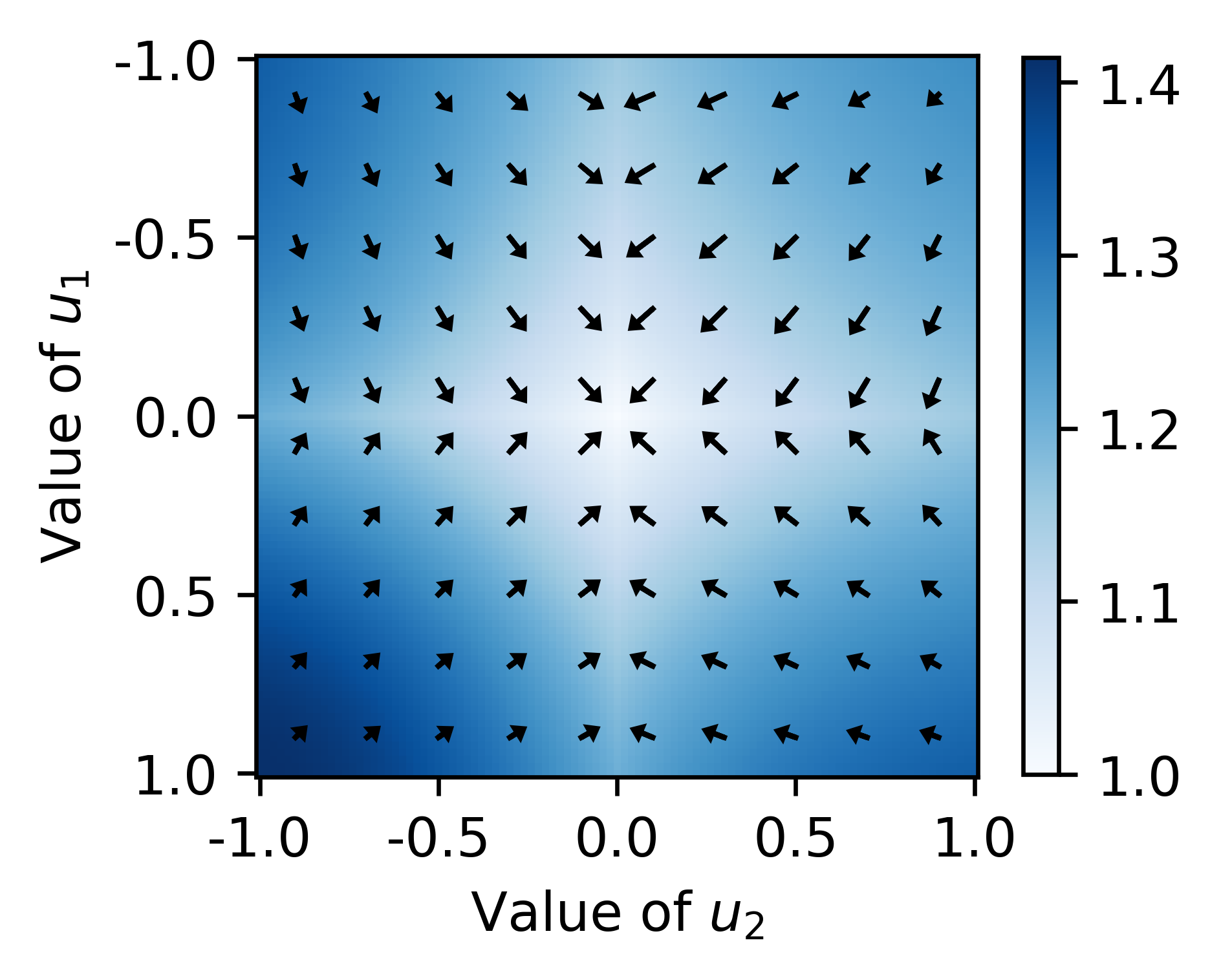}
    \vspace{-1mm}
    \caption{
        The value of $\valLoss$ (as color) for different values of $u_1$ and $u_2$, and the negative gradients of $\valLoss$ with respect to $u_1$ and $u_2$ (as arrows).
        $\valLoss$ achieves its minimum if $u_1 = u_2 = 0$, as claimed in Lemma \ref{lemma:general-case}, and the negative gradients point the optimum, supporting gradient-based optimization.
    }
    \label{fig:val-loss-proof}
\end{figure}

\myParagraph{Mean distance loss}
For ratio robustness, we propose to use the asymmetric mean distance between two sets as the $\mathrm{dist}$ function to separate $\mathrm{dist}(\trnEmb \cup \augEmb, \testEmb)$ into two terms:
\begin{equation}
    \valLoss(\trnEmb, \augEmb, \testEmb) =
    \frac{1}{2} {\textstyle \sum_{\mathbf{z}' \in \testEmb'}} \| \mathbf{z}' - \mu_\mathrm{trn}' \|_2 + \| \mathbf{z}' - \mu_\mathrm{aug}' \|_2 \;,
\label{eq:val-loss}
\end{equation}
where $\trnEmb'$, $\augEmb'$, and $\testEmb'$ are the embeddings after applying the total distance normalization, and $\mu_\mathrm{trn}'$ and $\mu_\mathrm{aug}'$ are the elementwise mean of $\trnEmb'$ and $\augEmb'$, respectively.
Computing the distances based on their averages allows $\valLoss$ to be invariant to the individual (or internal) distributions of $\trnEmb$ and $\augEmb$, including their sizes, while focusing on their global relative positions with respect to $\testEmb$.
In this way, we can avoid another trivial solution, which is to minimize $\valLoss$ only by decreasing the variance of $\augEmb$. 

\begin{figure*}
    \captionsetup[subfigure]{aboveskip=-0.5pt}
    \centering
    \includegraphics[width=0.75\textwidth]{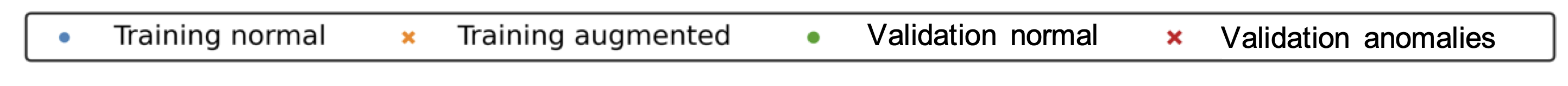}
    
    \begin{subfigure}{0.47\textwidth}
        \includegraphics[width=\textwidth]{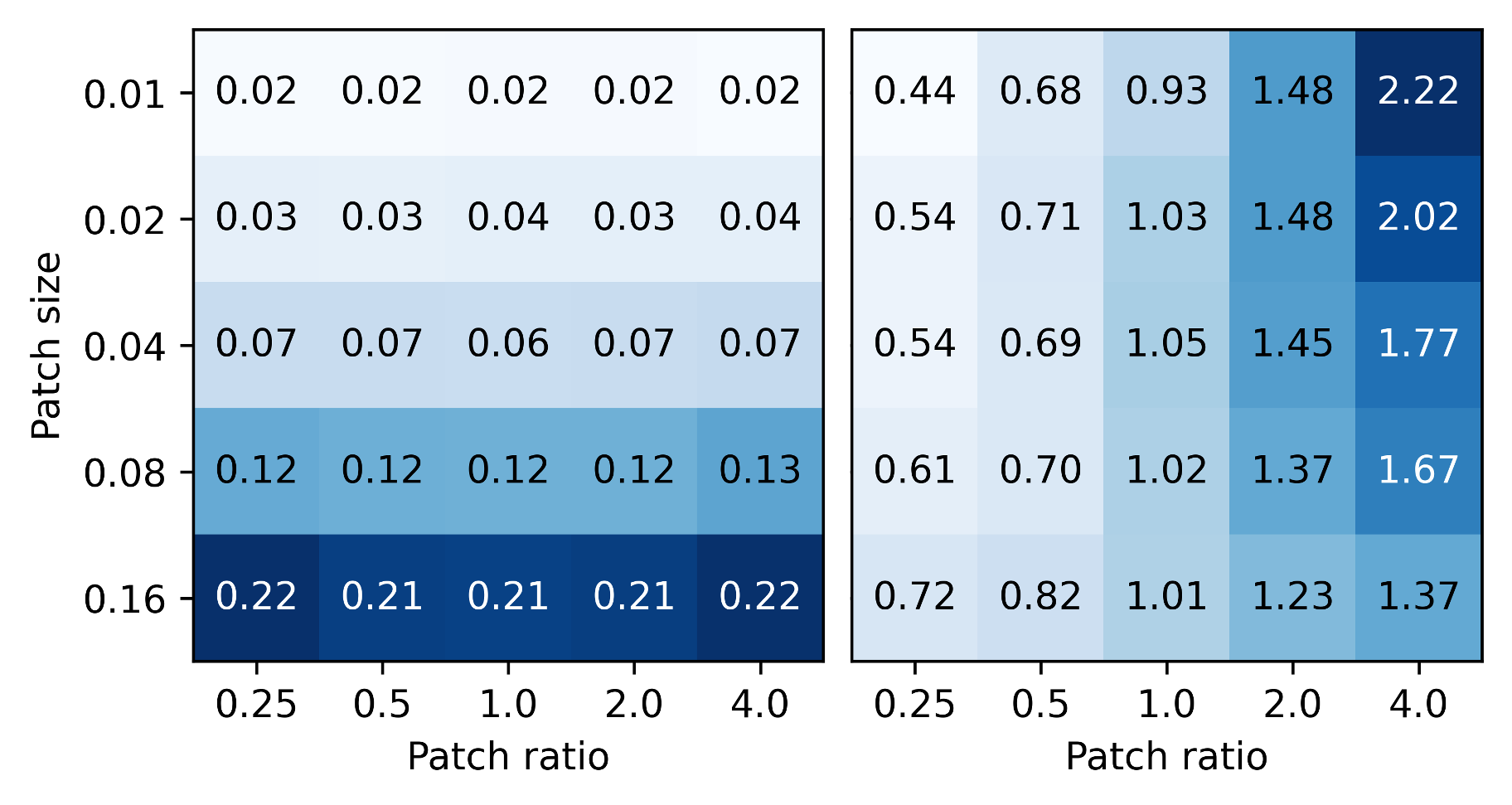}
        \caption{Learned values of patch size (left) \& ratio (right)}
        \label{fig:synthetic-exp-1}
    \end{subfigure} \hfill
    \begin{subfigure}{0.23\textwidth}
        \includegraphics[width=\textwidth]{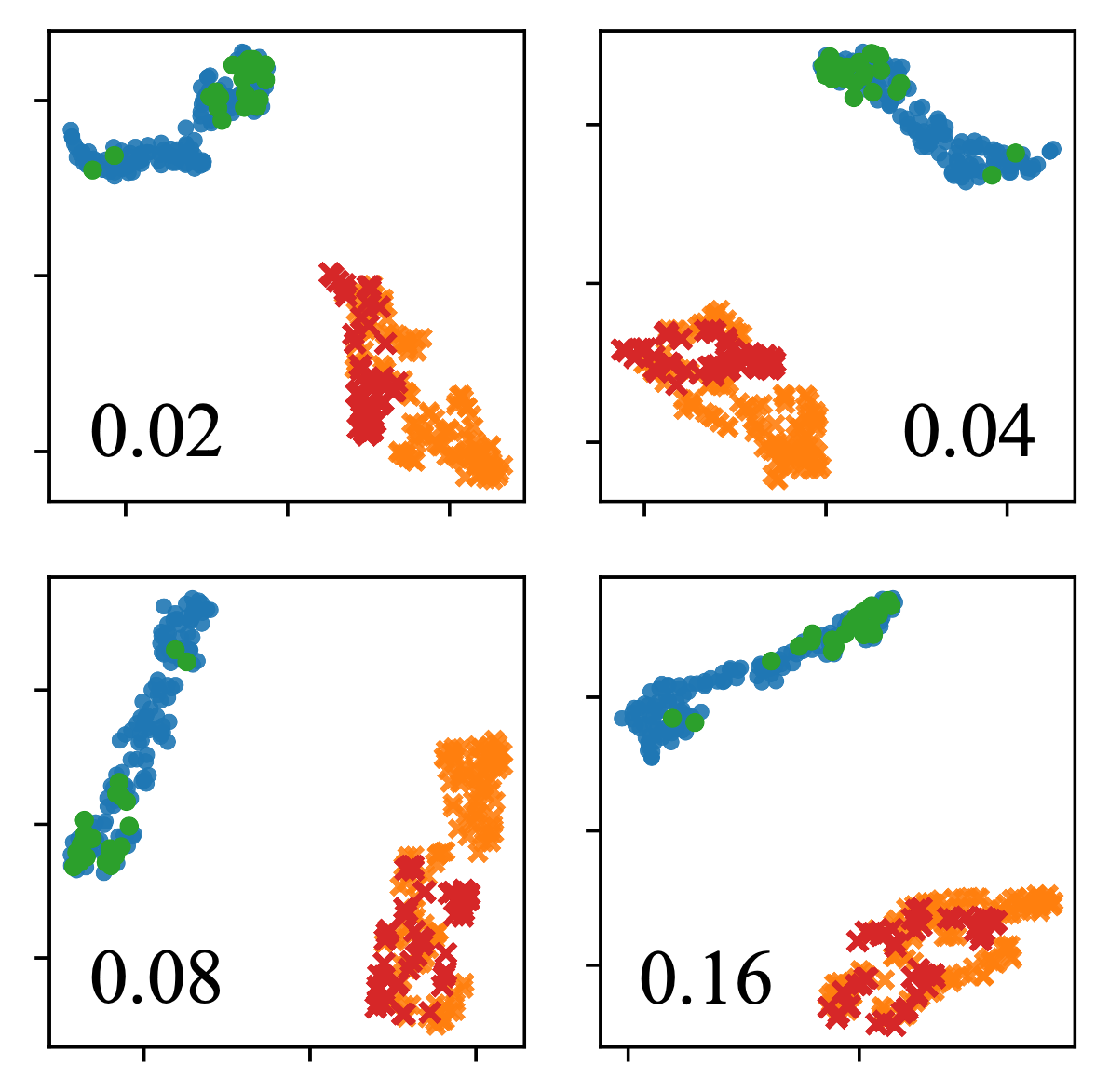}

        \vspace{3mm}
        \caption{$t$-SNE embeddings}
        \label{fig:synthetic-exp-2}
    \end{subfigure}
    \begin{subfigure}{0.28\textwidth}
        \includegraphics[width=\textwidth]{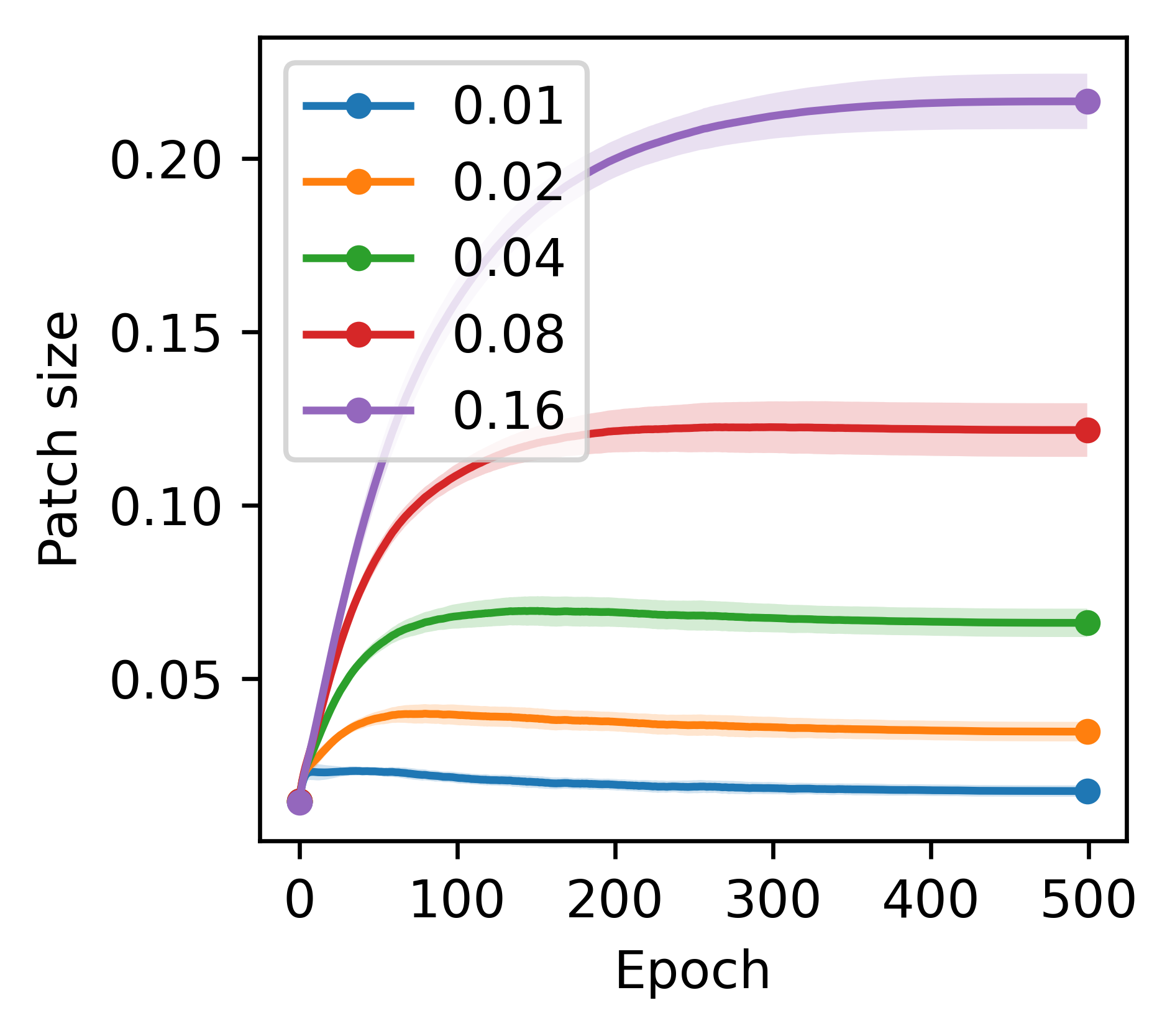}
        \caption{Patch size during training}
        \label{fig:synthetic-exp-3}
    \end{subfigure}

    \vspace{1mm}
    \caption{
        Experimental results on demonstrative examples, where we create 25 types of anomalies using CutDiff with five different patch size and patch ratio values, respectively.
        (a) Each cell's value represents what \method has learned through end-to-end optimization for the (left) patch size and (right) ratio, respectively.
        \method successfully mimics the true values of the patch ratio and size shown in the $x$- and $y$-axes, respectively, until (b) the embedding distributions are matched between $\trnEmb \cup \augEmb$ and $\testEmb$, showing a learning trajectory like (c).
        Note that AUC is $1.00$ in all 25 tasks.
    }
    \label{fig:synthetic-exp}
\end{figure*}

\myParagraph{Theoretical properties}
We give a theoretical analysis of $\valLoss$ on the singleton scenario, where the sets $\trnEmb$, $\augEmb$, \smash{$\testEmbN$}, and \smash{$\testEmbA$} all have size one.
Lemma \ref{lemma:perfect-alignment} shows that $\valLoss=1$ when the perfect alignment is achieved, and Lemma \ref{lemma:general-case} shows that we can actually reach the perfect alignment by minimizing $\valLoss$ via gradient-based optimization.
Fig. \ref{fig:val-loss-proof} provides empirical evidence on Lemma~\ref{lemma:general-case} when we assume that every embedding is a vector of length one.
All the negative gradients point to the perfect alignment $u_1 = u_2 = 0$.

\begin{lemma}[Perfect alignment]
    Assume $|\trnEmb| = |\augEmb| = 1$, \smash{$\trnEmb = \testEmbN$,} and \smash{$\augEmb = \testEmbA$}. Then, $\valLoss(\trnEmb, \augEmb, \testEmb) = 1$.
\label{lemma:perfect-alignment}
\end{lemma}

\begin{proof}
    The proof is given in Appendix \ref{appendix:proof-1}.
\end{proof}

\begin{lemma}[Existence of local optima]
    Let $\trnEmb = \{ \mathbf{z}_1 \}$, $\augEmb = \{ \mathbf{z}_2 \}$, \smash{$\testEmbN = \{ \mathbf{z}_1 + \mathbf{u}_1\}$}, and \smash{$\testEmbA = \{ \mathbf{z}_2 + \mathbf{u}_2\}$}.
    There exists $\delta > 0$ such that if $\| \mathbf{u}_1 \|_2 \leq \delta$ and $\| \mathbf{u}_2 \|_2 \leq \delta$, $\valLoss(\trnEmb, \augEmb, \testEmb) = 1$ if and only if $\mathbf{u}_1 = \mathbf{u}_2 = 0$.
\label{lemma:general-case}
\end{lemma}

\begin{proof}
    The proof is given in Appendix \ref{appendix:proof-2}.
\end{proof}

%% file: 042diffaug.tex
\subsection{Differentiable Augmentation}
\label{ssec:diffaug}

\begin{algorithm}[!t]
    \caption{Differentiable CutDiff Augmentation}
    \begin{algorithmic}[1]
        \REQUIRE
        Image $\mathbf{x}$ and augmentation hyperparameters $\mathbf{a} \in \mathbb{R}^3$
        \ENSURE
        Augmented image $\tilde{\mathbf{x}}$
        \STATE $\mathbf{X} \leftarrow$ Reshape $\mathbf{x}$ into size $m \times m \times 3$, assuming RGB
        \STATE $\mathbf{L} \leftarrow$ Reshape $\mathbf{a}$ as a lower triangular matrix of size $2 \times 2$
        \STATE $\bm{\mu} \leftarrow$ Sample a length-2 vector by $\mu_1, \mu_2 \sim \mathrm{uniform}(0, 1)$
        \STATE $\mathbf{g}_{ij} \leftarrow (i / m, j / m) \ \forall i, j \in [1, m]$
        \STATE $p_{ij} \leftarrow \exp(-(\mathbf{g}_{ij} - \bm{\mu})^\top (\mathbf{L} \mathbf{L}^\top)^{-1} (\mathbf{g}_{ij}  - \bm{\mu})) \ \forall i, j$
        \STATE \smash{$\tilde{x}_{ijk} \leftarrow \min(\max(x_{ijk} - p_{ij}, 0), 1) \ \forall i, j, k$}
        \STATE \smash{$\tilde{\mathbf{x}} \leftarrow$ Stack $\tilde{x}_{ijk}$ and reshape into the shape of $\mathbf{x}$}
  \end{algorithmic}
  \label{alg:cutdiff}
\end{algorithm}

The second component that enables \method to conduct end-to-end optimization is a differentiable augmentation function.
There are two main approaches to making augmentation differentiable.
The first is to train a neural network that mimics the augmentation function.
However, such a network is required to have high capacity to be able to learn the augmentation function accurately, especially for high-dimensional data such as images.
The second approach is to directly formulate an augmentation function in a differentiable way.
With a proper implementation or approximation, this approach can be more reliable and effective than the first approach.

We take the second approach and introduce two representative augmentation functions for local (or subtle) and gross (or semantic) anomalies, respectively, as a proof of concept.
Specifically, we propose CutDiff (\S\ref{ssec:cutdiff}) as a differentiable variant of CutOut \citep{Devries17CutOut}, which is a simple yet effective augmentation method for simulating \textit{local}ized anomalies.
We also utilize a differentiable formulation of
Rotation (\S\ref{ssec:rotation}), which in comparison transforms the input \textit{global}ly and has been widely used for semantic anomaly detection \citep{Bergman20GOAD, Sehwag21SSD}.

\subsubsection{CutDiff for Local Augmentation}
\label{ssec:cutdiff}

Local augmentation functions \citep{Devries17CutOut, Li21CutPaste, Schlter2021SelfSupervisedOD} such as CutOut and CutPaste mimic subtle local anomalies by modifying an image partially without changing the rest.
However, all of these functions are not differentiable, and thus cannot be used for our end-to-end optimization.

We propose CutDiff, a new differentiable variant of CutOut.
Note that most augmentation on images is a function of pixel positions, rather than pixel values; although the output is determined by the given image, the pixel values do not affect how the augmentation performs.
The main idea of CutDiff is to introduce a grid $\mathbf{G}$ of pixel locations as in Line 4 of Alg. \ref{alg:cutdiff} and design a differentiable function that takes $\mathbf{G}$ as an input and determines how to augment each pixel location based on the augmentation hyperparameters $\augParam$.

The process consists of three steps.
First, we sample the center position $\bm{\mu}$ as in Line 3 of Alg. \ref{alg:cutdiff}, which is a constant with respect to $\augParam$.
Then, for each image position $(i, j)$, we determine the amount of change to make with augmentation based on the distance from $\bm{\mu}$ scaled by $\augParam$ (in Line 5); the amount is smaller if $(i, j)$ is farther from $\bm{\mu}$.
Lastly, we replace the assignment (or replacement) operation in CutOut (and CutPaste) with the subtraction (in Line 6).
The $\mathrm{min}$ and $\mathrm{max}$ operations ensure that the output values are in $[0, 1]$.
Thanks to the subtraction, the information of the original pixels, not only their locations, is passed to the output image, allowing the gradient to flow to both the given image $\mathbf{x}$ and $\augParam$.

The three elements in $\augParam \in \mathbb{R}^3$ can be directly associated with the patch shape, making $\augParam$ interpretable and controllable.
We create a lower triangular matrix $\mathbf{L} \in \mathbb{R}^{2 \times 2}$ by reshaping $\augParam$ (shown in Line 2), which can be decomposed into $\mathbf{L} = \mathbf{R} \mathbf{S}$ with the rotation and scale matrices 
     $\mathbf{R} = \left[ \begin{matrix}
         \cos(g) & -\sin(g) \\
         \sin(g) & \cos(g)
     \end{matrix} \right] { \text{and} \; }
     \mathbf{S} = \left[ \begin{matrix}
         s / r & 0 \\
         0 & sr
     \end{matrix} \right]$,
where $g$, $s$, and $r$ represent the rotated angle, size, and ratio, respectively.
This means the role of $\augParam$ can be described in terms of the variables $g$, $s$, and $r$.
In Appendix~\ref{appendix:cutdiff}, we make visualization of images generated by CutDiff compared with CutOut and CutPaste.

It is noteworthy that the randomness of sampled $\bm{\mu}$ does not harm the gradient-based optimization of \method, since the validation loss $\valLoss$ is collectively computed on a set of embeddings, rather than on each individual sample unlike typical losses.

\subsubsection{Rotation for Global Augmentation}
\label{ssec:rotation}

Geometric augmentation functions such as rotation and translation have been widely used for image anomaly detection \citep{Golan18GEOM, Bergman20GOAD}.
Unlike local augmentation, many geometric transformations are differentiable if they are represented correctly as matrix-vector operations.
We adopt the differentiable rotation function proposed by \citet{Jaderberg15STN}, which consists of the two main stages.
The first stage is to create a $2 \times 3$ rotation matrix from a rotation angle, which is the same as the $\mathbf{R}$ matrix in CutDiff except that zeros are padded as the third column to match the shape.
The second stage is to create a sampling function that selects a proper pixel position of the given image for each position of the target image based on the rotation matrix and the affine grid.
The resulting operation is differentiable, since it is a parameterized mapping from the original pixels to the output pixels.

\begin{table*}
    \centering
    \caption{
        Test AUC on 23 different tasks for subtle anomaly detection.
        Each number is the average from five runs, and the best is in bold.
        \method outperforms most baselines, which is supported by the $p$-values derived by the Wilcoxon signed rank test.
    }
    \resizebox{0.99\textwidth}{!}{
    \begin{tabular}{l|l|cc|cccccc|c||ccc}
    \toprule
    \multicolumn{11}{c||}{Main Result} & \multicolumn{3}{c}{Ablation Study} \\
        \midrule
        Object & Anomaly Type & AE & D-SVDD & RS-CO & RD-CO & RS-CP & RD-CP & RS-CD & RD-CD & \method & MMD1 & MMD2 & FO \\
        \midrule
        Cable & Bent wire & 0.515 & 0.432 & 0.556 & 0.560 & 0.703 & \textbf{0.756} & 0.527 & 0.580 & 0.490 & 0.581 & 0.643 & 0.579 \\
        Cable & Cable swap & 0.639 & 0.295 & 0.483 & 0.625 & 0.618 & 0.683 & 0.574 & \textbf{0.696} & 0.532 & 0.510 & 0.562 & 0.545 \\
        Cable & Combined & 0.584 & 0.587 & 0.879 & 0.857 & 0.880 & \textbf{0.949} & 0.901 & 0.879 & 0.925 & 0.939 & 0.962 & 0.882 \\
        Cable & Cut inner insulation & 0.758 & 0.591 & 0.630 & 0.737 & 0.766 & \textbf{0.833} & 0.623 & 0.732 & 0.667 & 0.633 & 0.649 & 0.689 \\
        Cable & Cut outer insulation & \textbf{0.989} & 0.343 & 0.695 & 0.815 & 0.787 & 0.871 & 0.703 & 0.790 & 0.516 & 0.428 & 0.461 & 0.527 \\
        Cable & Missing cable & 0.920 & 0.466 & 0.953 & 0.961 & 0.755 & 0.801 & 0.935 & 0.945 & \textbf{0.998} & 0.855 & 0.772 & 0.999 \\
        Cable & Missing wire & 0.433 & 0.494 & 0.781 & 0.655 & 0.501 & 0.546 & 0.708 & 0.620 & \textbf{0.863} & 0.547 & 0.477 & 0.699 \\
        Cable & Poke insulation & 0.287 & 0.471 & 0.469 & 0.527 & 0.645 & \textbf{0.672} & 0.489 & 0.503 & 0.630 & 0.692 & 0.816 & 0.676 \\
        \midrule
        Carpet & Color & 0.578 & 0.716 & 0.669 & 0.508 & 0.412 & 0.287 & 0.643 & 0.639 & \textbf{0.938} & 0.761 & 0.741 & 0.918 \\
        Carpet & Cut & 0.198 & 0.758 & 0.439 & 0.608 & 0.403 & 0.411 & 0.490 & 0.767 & \textbf{0.790} & 0.353 & 0.401 & 0.595 \\
        Carpet & Hole & 0.626 & 0.676 & 0.379 & 0.613 & 0.404 & 0.389 & 0.470 & \textbf{0.765} & 0.590 & 0.438 & 0.229 & 0.630 \\
        Carpet & Metal contamination & 0.056 & \textbf{0.739} & 0.198 & 0.304 & 0.240 & 0.167 & 0.255 & 0.474 & 0.076 & 0.392 & 0.134 & 0.392 \\
        Carpet & Thread & 0.394 & \textbf{0.742} & 0.494 & 0.585 & 0.469 & 0.517 & 0.508 & 0.679 & 0.483 & 0.492 & 0.541 & 0.642 \\
        \midrule
        Grid & Bent & \textbf{0.849} & 0.168 & 0.456 & 0.322 & 0.421 & 0.433 & 0.337 & 0.354 & 0.771 & 0.780 & 0.650 & 0.602 \\
        Grid & Broken & 0.806 & 0.183 & 0.397 & 0.312 & 0.487 & 0.502 & 0.340 & 0.392 & \textbf{0.869} & 0.845 & 0.887 & 0.884 \\
        Grid & Glue & 0.704 & 0.143 & 0.634 & 0.568 & 0.674 & 0.732 & 0.681 & 0.578 & \textbf{0.906} & 0.966 & 0.974 & 0.721 \\
        Grid & Metal contamination & 0.851 & 0.229 & 0.421 & 0.380 & 0.499 & 0.514 & 0.425 & 0.613 & \textbf{0.858} & 0.861 & 0.665 & 0.732 \\
        Grid & Thread & 0.583 & 0.209 & 0.612 & 0.494 & 0.500 & 0.549 & 0.654 & 0.611 & \textbf{0.973} & 0.962 & 0.969 & 0.964 \\
        \midrule
        Tile & Crack & 0.770 & 0.728 & 0.872 & 0.993 & 0.743 & 0.636 & 0.837 & \textbf{0.999} & 0.749 & 0.740 & 0.820 & 0.595 \\
        Tile & Glue strip & 0.697 & 0.509 & 0.693 & \textbf{0.836} & 0.665 & 0.700 & 0.675 & 0.831 & 0.767 & 0.585 & 0.649 & 0.561 \\
        Tile & Gray stroke & 0.637 & 0.785 & 0.845 & 0.642 & 0.583 & 0.657 & 0.856 & 0.802 & \textbf{0.974} & 0.653 & 0.706 & 0.973 \\
        Tile & Oil & 0.414 & 0.690 & 0.708 & 0.745 & 0.464 & 0.576 & 0.683 & \textbf{0.837} & 0.554 & 0.548 & 0.614 & 0.555 \\
        Tile & Rough & 0.724 & 0.387 & 0.606 & \textbf{0.725} & 0.631 & 0.661 & 0.568 & 0.657 & 0.690 & 0.700 & 0.549 & 0.605 \\
        \midrule
        \multicolumn{2}{c|}{$p$-value} & \cellSecond{\cellFirst{.0000}} & \cellSecond{\cellFirst{.0000}} & \cellSecond{\cellFirst{.0000}} & \cellSecond{.0012} & \cellSecond{\cellFirst{.0000}} & \cellSecond{\cellFirst{.0000}} & \cellSecond{\cellFirst{.0000}} & .0728 & \textbf{Ours} & \cellThird{.0268} & \cellSecond{.0073} & .1332 \\    \bottomrule
    \end{tabular}
    }
    \label{tab:local-accuracy}
\end{table*}

%% file: 043impl.tex
\subsection{Techniques for Practical Usability}
\label{ssec:implement}

Following the main components, we introduce two additional techniques for improving the practical usability of \method for real-world data: \emph{second-order optimization} and \emph{multiple initialization}.

\myParagraph{Second-order optimization}
At each iteration, \method updates the hyperparameters $\augParam$ and the detector parameters $\theta$ through alternating stages, expecting the following to hold:
\begin{equation}
    \valLoss(\augParam^{(t+1)}, \theta') \;<\; \valLoss(\augParam^{(t)}, \theta) \;,
\label{eq:inequality}
\end{equation}
where $t$ is the current iteration, and $\theta'$ denotes the updated parameters of the detector network $\detector$ derived by using $\smash{\augParam^{(t)}}$ to generate pseudo anomalies $\augData$ for its training.

However, the first-order optimization of $\augParam$ cannot take into account that the parameters $\theta'$ and $\theta$ are different between both sides of Eq. \eqref{eq:inequality}, since it treats $\theta'$ as a constant.
As a solution, we consider $\theta'$ as a function of \smash{$\augParam^{(t)}$} and conducts second-order optimization:
\begin{equation}
    \augParam^{(t+1)} = \\ \augParam^{(t)} - \beta \nabla_{\augParam^{(t)}} \valLoss(\augParam^{(t)}, \theta - \alpha \nabla_\theta \trnLoss(\theta, \augParam^{(t)})) \;.
\label{eq:second-order}
\end{equation}
In this way, we can accurately track the change in $\theta$ caused by the update of $\augParam$, resulting in a stable minimization of $\valLoss$.
Note Eq.~\eqref{eq:second-order} is the same as in Line 8 of Algorithm \ref{alg:framework}, except that we use a slightly different notation for a better understanding.

\myParagraph{Multiple initialization}
The result of \method is also affected by how we initialize the augmentation hyperparameters $\augParam$, since it performs gradient-based updates toward a local optimum.
A natural way to address initialization  is to pick a few random starting points and select the best one.
However, it is difficult to select the best from multiple points, since $\valLoss$ is designed to locally improve the given $\augParam$, rather than to compare different models; e.g., it is possible that a less-aligned model produces lower $\valLoss$ if the augmented data are distributed more sparsely in the embedding space.

As a solution, we propose a simple yet effective way to compare models from different initialization choices.
Let $s$ be the anomaly score function presented in the problem definition in Sec. \ref{sec:prelim}.
Then, we use the score variance $\mathcal{S}$ to select the best initialization point after training completes (the larger, the better).
\begin{equation}
    \mathcal{S}(\theta) = \mathrm{variance}(\{ s(\detector(\mathbf{x})) \mid \mathbf{x} \in \testData \}).
\end{equation}
The intuition is that the variance of validation anomaly scores is likely to be large under good augmentation, as it generally reflects a better separability between inliers and anomalies.
This offers a fair evaluation since \method does not observe the function $s$.

%% file: 044complexity.tex
\subsection{Time Complexity}
\label{ssec:complexity}

The time complexity of \method is linear with respect to the data size and the number of parameters in the detector model $\detector$ despite the second-order optimization.
Here we provide theoretical analysis, which is further supported by empirical evidence in Section \ref{ssec:scalability}.

\begin{lemma}[Time complexity]
    Let $|\theta|$ be the number of parameters in $\detector$, $|\augParam|$ be the number of augmentation hyperparameters in $\augFunc$, $T$ be the number of training iterations, and $|\mathcal{D}|$ be the overall data size including all of training, validation, and test data.
    Then, the time complexity of ST-SSAD (Algorithm \ref{alg:framework}) is $O(T|\mathcal{D}|(|\theta| + |\augParam|)$.
\label{lemma:complexity}
\end{lemma}

\begin{proof}
    The proof is given in Appendix \ref{appendix:proof-3}.
\end{proof}

%% file: 050experiments.tex
\section{Experiments}
\label{sec:exp}

\myParagraph{Datasets}
We conduct experiments on \numData different anomaly detection tasks, which include 23 subtle (local) anomalies in MVTec AD \citep{Bergmann19MVTecAD} and 18 semantic (gross) anomalies in SVHN datasets \citep{Netzer11SVHN}.
MVTec AD contains images of industrial objects, where the anomalies are local defects such as scratches.
We use four types of objects, Cable, Carpet, Grid, and Tile, and their all anomaly types in experiments.
SVHN is a digits image dataset from house numbers.
We use digits 2 and 6 as normal classes and treat the rest as anomalies, generating 18 tasks for all pairs of digits.
Our experimental setup is different from \citep{Li21CutPaste} that uses all anomaly types at once, since we aim to evaluate the performance on each anomaly type.

\begin{table*}
\begin{minipage}{0.67\textwidth}
    \centering
    \caption{
        Test AUC for semantic anomaly detection.
        \method outperforms all baselines with the $p$-values smaller than $0.0001$.
        Same format as in Table \ref{tab:local-accuracy}.
    }
    \resizebox{\textwidth}{!}{\begin{tabular}{c|c|cc|cc|c||ccc}
    \toprule
    \multicolumn{7}{c||}{Main Result} & \multicolumn{3}{c}{Ablation Study} \\
        \midrule
        Object & Anomaly & AE & D-SVDD & RS-RO & RD-RO & \method & MMD1 & MMD2 & FO \\
        \midrule
        Digit 2 & Digit 0 & 0.602 & 0.472 & 0.672 & 0.734 & \textbf{0.816} & 0.519 & 0.518 & 0.506 \\
        Digit 2 & Digit 1 & 0.544 & 0.499 & 0.601 & 0.609 & \textbf{0.743} & 0.499 & 0.501 & 0.498 \\
        Digit 2 & Digit 3 & 0.604 & 0.503 & 0.664 & 0.730 & \textbf{0.832} & 0.508 & 0.510 & 0.511 \\
        Digit 2 & Digit 4 & 0.561 & 0.511 & 0.679 & 0.746 & \textbf{0.790} & 0.531 & 0.530 & 0.530 \\
        Digit 2 & Digit 5 & 0.625 & 0.502 & 0.709 & 0.824 & \textbf{0.877} & 0.512 & 0.514 & 0.517 \\
        Digit 2 & Digit 6 & 0.616 & 0.496 & 0.726 & 0.826 & \textbf{0.887} & 0.511 & 0.507 & 0.510 \\
        Digit 2 & Digit 7 & 0.541 & 0.496 & 0.584 & 0.639 & \textbf{0.823} & 0.521 & 0.520 & 0.518 \\
        Digit 2 & Digit 8 & 0.616 & 0.498 & 0.673 & 0.738 & \textbf{0.805} & 0.524 & 0.524 & 0.522 \\
        Digit 2 & Digit 9 & 0.588 & 0.485 & 0.625 & \textbf{0.687} & 0.659 & 0.516 & 0.523 & 0.518 \\
        \midrule
        Digit 6 & Digit 0 & 0.531 & 0.480 & 0.586 & 0.606 & \textbf{0.777} & 0.503 & 0.514 & 0.504 \\
        Digit 6 & Digit 1 & 0.517 & 0.498 & 0.621 & 0.610 & \textbf{0.854} & 0.516 & 0.528 & 0.511 \\
        Digit 6 & Digit 2 & 0.594 & 0.503 & 0.735 & 0.807 & \textbf{0.916} & 0.525 & 0.531 & 0.529 \\
        Digit 6 & Digit 3 & 0.570 & 0.507 & 0.686 & 0.750 & \textbf{0.823} & 0.518 & 0.520 & 0.521 \\
        Digit 6 & Digit 4 & 0.525 & 0.508 & 0.659 & 0.703 & \textbf{0.709} & 0.527 & 0.521 & 0.530 \\
        Digit 6 & Digit 5 & 0.544 & 0.502 & 0.615 & \textbf{0.661} & 0.658 & 0.516 & 0.513 & 0.509 \\
        Digit 6 & Digit 7 & 0.567 & 0.505 & 0.699 & 0.729 & \textbf{0.861} & 0.540 & 0.551 & 0.541 \\
        Digit 6 & Digit 8 & 0.546 & 0.500 & 0.575 & 0.641 & \textbf{0.732} & 0.512 & 0.523 & 0.512 \\
        Digit 6 & Digit 9 & 0.579 & 0.495 & 0.708 & 0.817 & \textbf{0.944} & 0.527 & 0.519 & 0.524 \\
        \midrule
        \multicolumn{2}{c|}{$p$-value} & \cellSecond{\cellFirst{.0000}} & \cellSecond{\cellFirst{.0000}} & \cellSecond{\cellFirst{.0000}} & \cellSecond{\cellFirst{.0000}} & \textbf{Ours} & \cellSecond{\cellFirst{.0000}} & \cellSecond{\cellFirst{.0000}} & \cellSecond{\cellFirst{.0000}} \\    \bottomrule
    \end{tabular}
    }
    \label{tab:gross-accuracy}
\end{minipage} \hfill
\begin{minipage}{0.31\textwidth}
    \centering
    \vspace{-1mm}
    \includegraphics[width=0.9\textwidth]{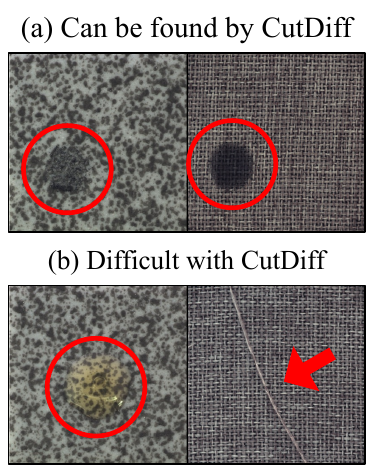}
    \captionof{figure}{
        Some anomalies are hard to detect by CutDiff due to the inherent mismatch between $\augFunc$ and the anomaly-generating mechanism.
        See Sec. \ref{ssec:discuss} for detailed discussion.
    }
    \label{fig:limitation}
\end{minipage}
\end{table*}

\myParagraph{Model setting}
Our model setting follows a previous work \citep{Li21CutPaste}, including the ResNet-based detector $\detector$ and the cross entropy training loss $\trnLoss$.
As the anomaly score function $s$, we use the negative log likelihood of a Gaussian density estimator learned on the embeddings of training data \citep{Rippel20GDE, Li21CutPaste}.
For \method, we use four initialization points for each augmentation function: $\{0.0001, 0.001, 0.01, 0.1\}$ for the CutDiff patch size, and $\{45^\circ, 135^\circ, 225^\circ, 315^\circ\}$ for the Rotation angle.
We set both the initial patch angle and ratio to zero.
We employ CutDiff and Rotation for subtle and semantic anomaly detection, respectively.
The sum of $\trnLoss$ and $\valLoss$ is used as the stopping criterion for hyperparameter updates.

\myParagraph{Evaluation metrics}
The performance of each model is measured by the area under the ROC curve (AUC) on anomaly scores computed for test data.
We run all experiments five times and report the average as the main result.
For statistical comparison between different models, we also run the paired Wilcoxon signed-rank test \citep{Groggel00Stats}.
The one-sided test with $p$-values less than $0.05$ represents that our \method is statistically better than the other.

\myParagraph{Baselines}
To the best of our knowledge, there are no competitors on end-to-end augmentation hyperparameter tuning for SSAD.
Thus, we compared \method with various types of baselines:
\emph{SSL without hyperparameter tuning}---(1) random dynamic selection (RD) that selects $\augParam$ randomly at each training epoch, and (2) random static selection (RS) that selects $\augParam$ once before the training begins.
\emph{Unsupervised learning}---(3) autoencoder (AE) \citep{Golan18GEOM} and (4) DeepSVDD \citep{Ruff18DeepSVDD}.
\emph{Variants of our \method with na\"ive choices}---(5) using maximum mean discrepancy (MMD) \citep{Gretton06MMD} as $\valLoss$, (6) MMD without the total distance normalization, and (7) using first-order optimization.
RD and RS are used with either CutOut, CutPaste, CutDiff, or Rotation, which we denote CO, CP, CD, and RO, respectively, for brevity.
We also denote baselines (5)--(7) as MMD1, MMD2, and FO, respectively.
Additional details on experimental settings can be found in Appendix \ref{appendix:exp}.

\subsection{Demonstrative Examples}
\label{ssec:exp-demo}

We present experimental results on demonstrative datasets, where we create synthetic anomalies through CutDiff.
Given normal images of the Carpet object, we generate 25 types of anomalies with the patch size from $0.01$ to $0.16$ and the aspect ratio from $0.25$ to $4.0$, where the angle is fixed to $0$.
Our goal is to demonstrate that \method is able to learn different values of $\augParam$ for different anomaly types even when the normal training data are fixed.

Fig. \ref{fig:synthetic-exp} summarizes the results.
In Fig. \ref{fig:synthetic-exp-1}, \method learns different values of $\augParam$ depending on the anomaly types, specifically patch size and patch ratio, demonstrating the ability of \method to adapt to varying anomalies.
There is a slight difference between the learned $\augParam$ and the true values in some cases, since embedding distributions between $\trnData \cup \augData$ and $\testData$ can be matched as shown in Fig. \ref{fig:synthetic-exp-2} even with a small difference; the training of \method stops after the distributions are matched.
This difference is typically larger for patch ratio than for patch size, suggesting that patch size impacts the embeddings more than the ratio does.

Fig. \ref{fig:synthetic-exp-3} depicts the training process of \method for five anomaly types generated with different patch sizes, where the patch ratio is set to $1.0$.
We visualize the average and standard deviation from five runs with different random seeds.
\method accurately adapts to different patch sizes even from the same initialization, updating $\augParam$ through iterations to minimize the validation loss.

\begin{figure*}
    \centering
    \includegraphics[width=0.92\textwidth]{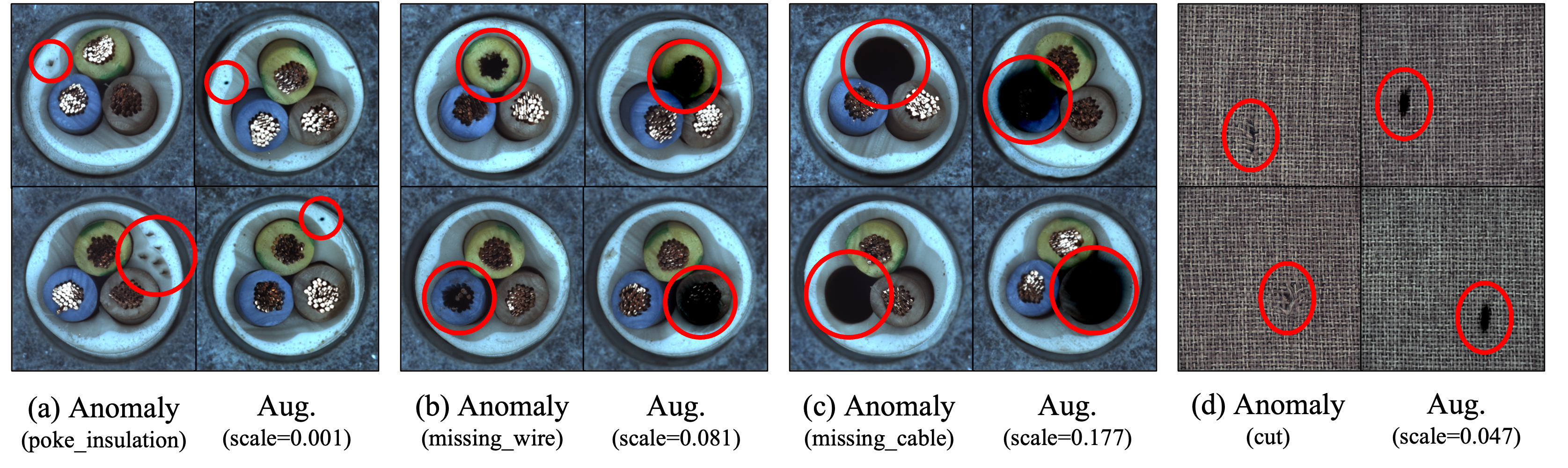}
    \vspace{-1mm}
    \caption{
        Illustrations of four anomaly types for the Cable and Carpet objects and the augmentations learned by \method.
        Different hyperparameters of CutDiff are learned to resemble the true anomalies, including the size and the aspect ratio.
    }
    \label{fig:local-images}
\end{figure*}

\subsection{Testbed Evaluation}
\label{ssec:quant}

\myParagraph{Quantitative analysis}
Next, we perform quantitative evaluation of \method on both subtle and semantic class anomalies, covering 41 different anomaly detection tasks.
Table \ref{tab:local-accuracy} provides the results on 23 tasks for industrial-defect anomalies.
\method achieves the best AUC in 9 different tasks, and outperforms 7 of the 8 baselines with $p$-values smaller than $0.01$, showing strong statistical significance in all 23 tasks.
The $p$-value is still small for the remaining case ($0.07$) when compared to RD-CD, which means that \method outperforms RD-CD in most tasks in the testbed.
Table \ref{tab:gross-accuracy} shows the results on 18 tasks for semantic class anomalies, where \method significantly outperforms all baselines with all $p$-values smaller than $0.0001$.

The ablation studies between \method and its variants in Tables \ref{tab:local-accuracy} and \ref{tab:gross-accuracy} show the effectiveness of our ideas that compose \method: total distance normalization, mean distance loss, and second-order optimization.
Especially, the two MMD-based methods are significantly worse than \method, showing that MMD is not suitable for augmentation tuning even though it is a popular distance measure, due to the challenges we address with $\valLoss$.
The difference between \method and its first-order variant is smaller, meaning that the first-order optimization can still be used for efficiency.


\myParagraph{Qualitative analysis}
We also perform a qualitative analysis, visualizing the augmentation functions learned by \method for different anomaly types.
Fig. \ref{fig:local-images} illustrates three types of anomalies in the Cable object and one type of anomaly in the Carpet object.
The four anomaly types have their own sizes and aspect ratios of defected regions, which are accurately learned by \method.
Note that the three Cable anomalies share the same training data, but \method captures their difference well from the alignment.
The patch locations are chosen randomly at each run, since the locations of local defects are different for each anomalous image.

\subsection{Discussion on Limitations}
\label{ssec:discuss}

Table \ref{tab:local-accuracy} shows the success of \method throughout different tasks, but also its limitation in some tasks.
In tasks like Rough anomalies in Tile, a simple baseline like random CutOut shows the best AUC.
This is because some anomaly types are hard to mimic with CutDiff due to the inherent mismatch of the augmentation function.

Fig.~\ref{fig:limitation} shows two example anomaly types, Tile-Oil and Carpet-Thread, where \method cannot improve over the baselines.
Local defects of Oil are brighter than the background, whereas CutDiff always darkens the chosen patch regardless of the hyperparameter values.
The anomalies of the Thread type contain long thin threads, which are hard to represent with CutDiff in nature.
Such a behavior is expected for \method, which relies on gradient-based updates to tune the augmentation hyperparameters $\augParam$; the space it can explore during the optimization is determined by the amount of changes that can be created by changing continuous hyperparameters.


\subsection{Linear Complexity}
\label{ssec:scalability}

\begin{figure}
    \begin{subfigure}{0.235\textwidth}
        \includegraphics[width=\textwidth]{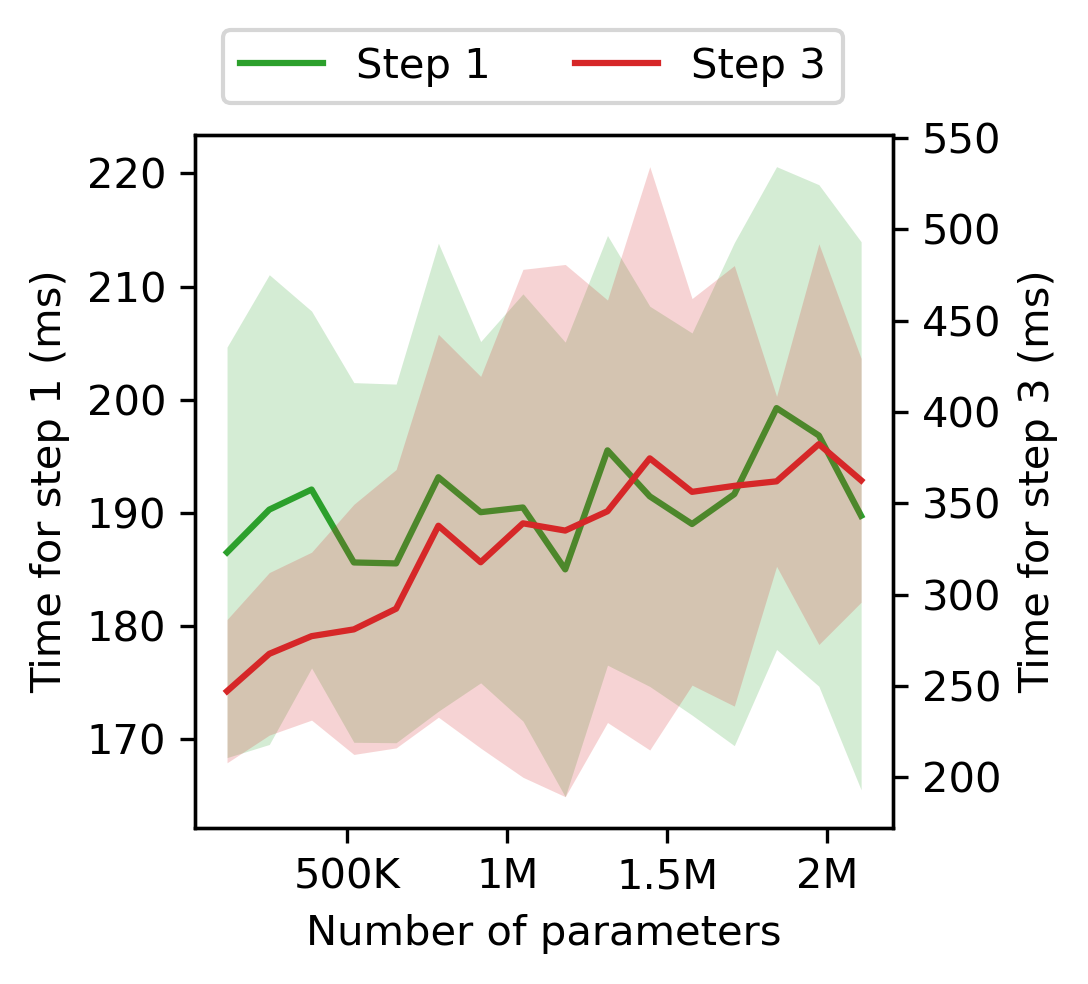}
        \vspace{-0.2in}
        \caption{Steps 1 and 3.}
        \label{fig:time-1}
    \end{subfigure} \hfill
    \begin{subfigure}{0.235\textwidth}
        \includegraphics[width=\textwidth]{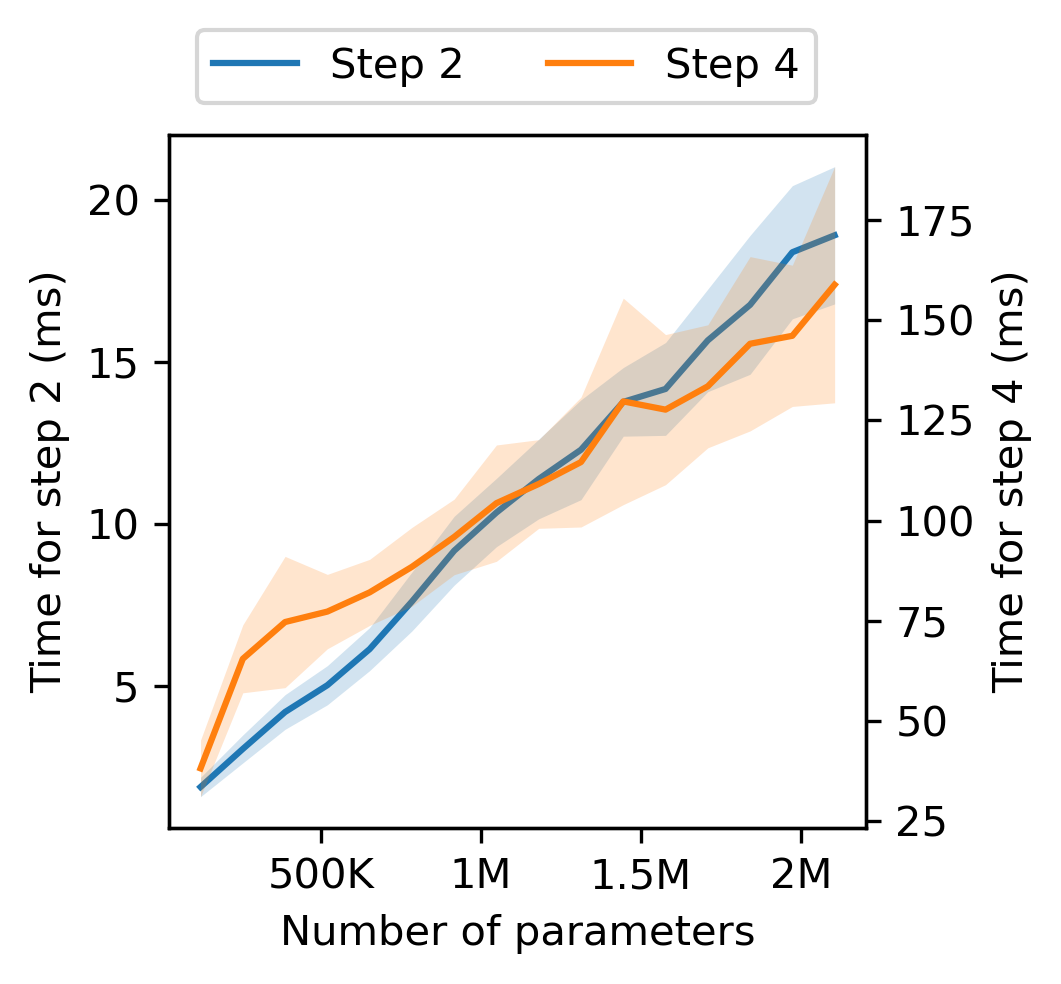}
        \vspace{-0.2in}
        \caption{Steps 2 and 4.}
        \label{fig:time-2}
    \end{subfigure}

    \vspace{2mm}
    \caption{Running time of each step of ST-SSAD. We report the average and standard deviation from 100 runs.}
    \label{fig:time}
\end{figure}

As supporting evidence for the time complexity, we run an additional experiment on ST-SSAD by varying the number of parameters in the detector $\detector$.
Here we use MLP-based architecture to create a direct relationship between  inference time and the number of parameters, while controlling other variables that might affect the experimental result.
We measure running time for each step of \method per training epoch, while changing the number of layers in $\detector$ from 8 to 128.
Fig. \ref{fig:time} presents the results, dividing the four steps into two plots.
Steps 1-4 correspond to the forward pass of Line 3, the backward pass of Line 3, Lines 4-7, and Line 8 of Algorithm \ref{alg:framework}, respectively, with the details given in Appendix \ref{appendix:proof-3}.

We have two main observations.
First, the running time increases sub-linearly with respect to model size, as expected from our theoretical analysis.
Second, running times of Steps 1, 3 and 4 are almost 10$\times$ longer than in Step 2, since they depend on the data size, while Step 2 does not.
These results demonstrate that \method is scalable despite the second-order optimization, as it does not require computing the Hessian matrix (details in Appendix \ref{appendix:proof-3}).

%% file: 060conclusion.tex
\section{Conclusion}
\label{sec:discuss}

We presented \method, the first framework for automatically tuning the augmentation hyperparameters of self-supervised anomaly detection (SSAD) in a systematic fashion.
Within our end-to-end framework, we addressed two key research challenges: unsupervised validation and differentiable augmentation.
The former proposed a smooth validation loss that quantifies the agreement between augmented data and unlabeled validation data.
The latter introduced two differentiable formulations for both local and global augmentation, while \method can flexibly accommodate any other differentiable augmentation.
Our experiments on large testbeds validated the superiority of \method over existing practices, as well as showcased its scalability.
\method is a general framework, rather than a specific method, and can accommodate other detectors and augmentation functions.
We leave it as future work to deal with more diverse types of anomalies and data modalities.

%% file: 100appendix.tex
\input{110proofs}
\input{120cutdiff}

\input{130expset}

%% file: 110proofs.tex
\section{Proofs of Lemmas}
\label{appendix:valloss}

\subsection{Proof of Lemma \ref{lemma:perfect-alignment}}
\label{appendix:proof-1}

\begin{proof}
Let $\trnEmb = \testEmbN = \{ \mathbf{z}_1 \}$ and $\augEmb = \testEmbA = \{ \mathbf{z}_2 \}$.
Then, the embedding matrix \smash{$\mathbf{Z} \in \mathbb{R}^{h \times 4}$} before the total distance normalization is given as $\mathbf{Z} = \left[ \begin{matrix}
        \mathbf{z}_1 &
        \mathbf{z}_1 &
        \mathbf{z}_2 &
        \mathbf{z}_2
    \end{matrix} \right]$.
Let $\bar{\mathbf{z}} = (\mathbf{z}_1 + \mathbf{z}_2) / 2$, and \smash{$\bar{\mathbf{Z}} \in \mathbb{R}^{4 \times h}$} be a matrix where each column is $\bar{\mathbf{z}}$.
Then, $\mathbf{Z}$ is transformed into \smash{$\tilde{\mathbf{Z}}$} as a result of the normalization:
\begin{align*}
    \tilde{\mathbf{Z}}
    &= \frac{2}{\sqrt{
        2\sum_i (z_{1i} - \bar{z}_i)^2 + 2 \sum_i (z_{2i} - \bar{z}_i)^2
    }}
    (\mathbf{Z} - \bar{\mathbf{Z}}) \\
    &= \frac{2}{\sqrt{\sum_i
        (z_{1i} - z_{2i})^2
    }}
    (\mathbf{Z} - \bar{\mathbf{Z}})
    = \frac{2}{
        \| \mathbf{z}_1 - \mathbf{z}_2 \|_2
    }
    (\mathbf{Z} - \bar{\mathbf{Z}}).
\end{align*}
The validation loss $\valLoss$ is computed on \smash{$\tilde{\mathbf{Z}}$} as follows:
\begin{align*}
    &\valLoss(\trnEmb, \augEmb, \testEmb) \\
    &= \frac{1}{4} \left(
        \| \tilde{\mathbf{z}}_1 - \tilde{\mathbf{z}}_1 \|_2 +
        \| \tilde{\mathbf{z}}_1 - \tilde{\mathbf{z}}_2 \|_2 +
        \| \tilde{\mathbf{z}}_2 - \tilde{\mathbf{z}}_1 \|_2 +
        \| \tilde{\mathbf{z}}_2 - \tilde{\mathbf{z}}_2 \|_2
    \right) \\
    &= \frac{1}{2} \| \tilde{\mathbf{z}}_1 - \tilde{\mathbf{z}}_2 \|_2 \\
    &= \frac{1}{2} \left\|
        \frac{2}{
            \| \mathbf{z}_1 - \mathbf{z}_2 \|_2
        } ((\mathbf{z}_1 - \bar{\mathbf{z}}) - (\mathbf{z}_2 - \bar{\mathbf{z}})) 
    \right\|_2 \\
    &= 1.
\end{align*}
As a result, the lemma is proved.
\end{proof}

\subsection{Proof of Lemma \ref{lemma:general-case}}
\label{appendix:proof-2}

\begin{proof}
Without loss of generality, we assume the simplest scalar embeddings of size one as $\trnEmb = \{ 0 \}$, $\testEmbN = \{ u_1 \}$, $\augEmb = \{ 2 \}$, $\testEmbA = \{ u_2 + 2 \}$.
Then, we show that $\valLoss(\trnEmb, \augEmb, \testEmb)=1$ if and only if $u_1 = u_2 = 0$ when $\delta = 1$.
First, we represent the embedding vector (which is used to be a matrix) \smash{$\mathbf{z} \in \mathbb{R}^4$} before the total distance normalization as $\mathbf{z} = (0, u_1, 2, u_2 + 2)$.

Let $\bar{z} = (u_1 + u_2 + 4) / 4$ be the center, and \smash{$\bar{\mathbf{z}} \in \mathbb{R}^4$} be a vector where each element is $\bar{z}$.
Then, $\mathbf{z}$ is transformed into $\tilde{\mathbf{z}}$ as a result of the normalization:
\begin{align*}
    \tilde{\mathbf{z}}
    &= \frac{2}{\sqrt{
        \bar{z}^2 + (u_1 - \bar{z})^2 + (2 - \bar{z})^2 + (u_2 + 2 - \bar{z})^2
    }} (\mathbf{z} - \bar{\mathbf{z}}) \\
    &= \frac{2}{\sqrt{
        4 \bar{z}^2 - 2(u_1 + u_2 + 4) \bar{z} + u_1^2 + u_2^2 + 4u_2 + 8
    }} (\mathbf{z} - \bar{\mathbf{z}}) \\
    &= \frac{2}{\sqrt{
        u_1^2 + u_2^2 + 4u_2 + 8 - 4 \bar{z}^2
    }} (\mathbf{z} - \bar{\mathbf{z}}) \\
    &= \frac{4}{\sqrt{
        3u_1^2 + 3u_2^2 - 8u_1 + 8u_2 - 2u_1u_2 + 16
    }} (\mathbf{z} - \bar{\mathbf{z}}).
\end{align*}
The validation loss $\valLoss$ is computed on $\tilde{\mathbf{z}}$ as follows:
\begin{align*}
    \valLoss(\trnEmb, \augEmb, \testEmb)
    &= \frac{|u_1| + |u_1 - 2| + |u_2| + |u_2 + 2|}{\sqrt{
        3u_1^2 + 3u_2^2 - 8u_1 + 8u_2 - 2u_1u_2 + 16
    }}.
\end{align*}
Then, we can consider four cases based on whether $u_1 \geq 0$ and $u_2 \geq 0$.
To show that the inequality $\valLoss(\trnEmb, \augEmb, \testEmb) \geq 1$ holds, we represent $\valLoss^2(\trnEmb, \augEmb, \testEmb) - 1$ as follows:
\begin{align*}
    -3u_1^2 + u_2^2 + 8u_1 + 8u_2 + 2u_1u_2
        & \textrm{ if } u_1 \geq 0 \textrm{ and } u_2 \geq 0 \\
    -3u_1^2 - 3u_2^2 + 8u_1 - 8u_2 + 2u_1u_2
        & \textrm{ if } u_1 \geq 0 \textrm{ and } u_2 \leq 0 \\
    u_1^2 + u_2^2 - 8u_1 + 8u_2 - 6u_1u_2
        & \textrm{ if } u_1 \leq 0 \textrm{ and } u_2 \geq 0 \\
    u_1^2 - 3u_2^2 - 8u_1 - 8u_2 + 2u_1u_2
        & \textrm{ if } u_1 \leq 0 \textrm{ and } u_2 \leq 0
\end{align*}
It is straightforward to see that $\valLoss^2(\trnEmb, \augEmb, \testEmb) - 1 \geq 0$ in all four cases if $-1 \leq u_1 \leq 1$ and $-1 \leq u_2 \leq 1$, and the equality holds if and only if $u_1 = u_2 = 0$.
Since $\valLoss(\trnEmb, \augEmb, \testEmb) \geq 0$ by its definition, we prove the lemma for the scalar case.
The extension to multi-dimensional cases is trivial, since all operations in $\valLoss$ can be generalized.
\end{proof}

\subsection{Proof of Lemma \ref{lemma:complexity}}
\label{appendix:proof-3}

\begin{proof}

We analyze the time complexity of \method based on Algorithm~\ref{alg:framework}.
Since \method is a training framework which runs for $T$ iterations, it is obvious that its time complexity is proportional to $T$.
We focus on how each iteration works.
\begin{itemize}
    \item \textbf{Step 1 (forward pass of Line 3):} $O(|\mathcal{B}|(|\theta| + |\augParam|)$.
        Let $\mathcal{B}$ be a mini-batch of data, where $|\mathcal{B}|$ is the batch size.
        During the forward pass, we run both $\phi_\mathrm{aug}$ and $f_\theta$ for $\mathcal{B}$ to compute the training loss $\trnLoss$ for the current parameters \smash{$\theta^{(t)}$ and $\augParam^{(t)}$}.
    \item \textbf{Step 2 (backward pass of Line 3):} $O(|\theta| + |\augParam|)$.
        During the backward pass, we get new detector parameters \smash{$\theta^{(t+1)}$} as a function of \smash{$\theta^{(t)}$} and \smash{$\augParam^{(t)}$} for the second-order optimization.
        The only difference from the typical first-order pass is that we store the backward computational chain as additional information, colored in orange in Fig. \ref{fig:backward}, where the size of the additional chain is $O(|\theta|)$.
        Thus, the overall complexity of the backward pass is linear with the model size.

\begin{figure}[H]
    \centering
    \includegraphics[width=0.45\textwidth, trim={1.5cm 0 1.5cm 0}, clip]{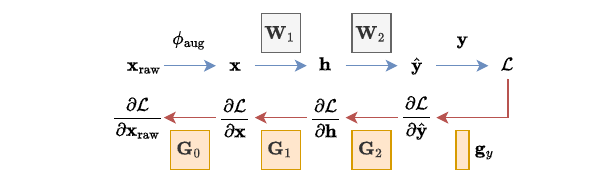}
    \caption{Illustration of the backward pass.}
    \label{fig:backward}
\end{figure}

    \item \textbf{Step 3 (Lines 4 - 7):} $O(|\mathcal{D}|(|\theta| + |\augParam|))$.
        This step is straightforward, since \method runs $\detector$ and $\augFunc$ for all data and normalize the embeddings, both of which run in linear time.
    \item \textbf{Step 4 (Line 8):} $O(|\mathcal{D}|(|\theta| + |\augParam|))$.
        We compute new augmentation hyperparameters $\augParam^{(t+1)}$ after backpropagation to the original ones \smash{$\augParam^{(t)}$}.
        This is slower than Step 2, since all the data explicitly participate in computing $\valLoss$.
\end{itemize}

In summary, the time complexity of ST-SSAD is $O(T|\mathcal{D}|(|\theta| + |\augParam|))$, where $T$ is the number of training iterations.
\end{proof}

%% file: 120cutdiff.tex
\section{Visualization of CutDiff}
\label{appendix:cutdiff}

\begin{figure*}[t]
    \centering
    \includegraphics[width=0.75\textwidth]{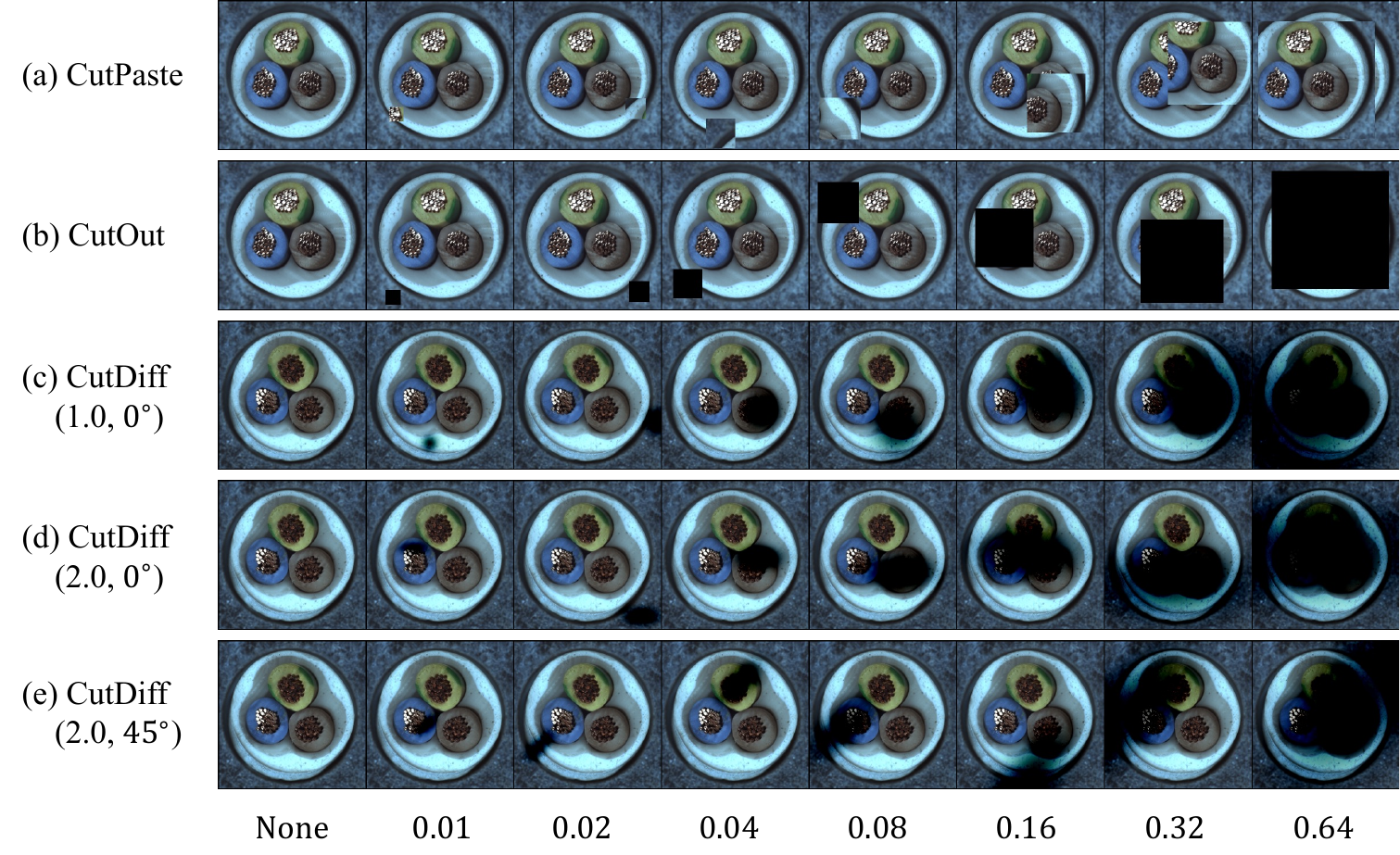}
    \caption{
        Comparison between (a) CutPaste, (b) CutOut, and (c - e) CutDiff with different hyperparameter choices.
        Each column represents a different patch size in $[0.01, 0.64]$, and the two numbers in (c - e) represent the aspect ratio and the rotated angle of a patch, respectively.
        CutDiff creates a smooth differentiable boundary unlike the other two functions.
    }
    \label{fig:cutdiff-images}
\end{figure*}


Fig. \ref{fig:cutdiff-images} shows the images generated from CutPaste, CutOut, and CutDiff, respectively, with different hyperparameter choices.
CutDiff is similar to CutOut, except that it creates a smooth circular patch which supports gradient-based updates of its hyperparameters.
It is clear that the two other augmentations, CutPaste and CutOut, are not differentiable as they replace the original pixels with new ones (either black for CutOut or a copied patch for CutPaste).

%% file: 130expset.tex
\section{Detailed Experimental Setup}
\label{appendix:exp}

\subsection{Implementation Details}

\method contains two implementation techniques which are not introduced in the main paper due to the lack of space.
The first technique is \emph{warm start,} which means that we train the detector $\detector$ for a fixed number of epochs before starting the alternating updates of $\theta$ and $\augParam$.
Such a warm start is required since the gradient-based update of $\augParam$ is ineffective if the detector $\detector$ does not perform well on the current $\augParam$; in such a case, the validation loss $\valLoss$ can have an arbitrary value which is not related to the true alignment of $\augParam$.
The number of training epochs for warm start is chosen to sufficiently minimize the training loss $\trnLoss$ for the initial $\augParam$.

The second implementation technique is to update $\theta$ multiple times for each update of $\augParam$.
This is based on the same motivation as in the first technique; we need to ensure the reasonable performance of $\detector$ during the iterative updates.
The default choice is only one update of $\theta$, but if the training loss $\trnLoss$ does not decrease enough at each iteration, one may consider increasing the number of updates.
That is, the number of updates can be determined by observing the decrease of $\trnLoss$ during training even without labeled data.

\subsection{Hyperparameter Choices}

We choose the model and training hyperparameters of \method based on previous work \citep{Li21CutPaste}, including the choice of the detector network $\detector$ and the score function $s$, and use them across all tasks in our experiments for consistency with previous results.
On the other hand, we tune some of the hyperparameters that need to be controlled based on the properties of each dataset:
\begin{compactitem}
    \item Batch size: 32 (in MVTecAD) and 256 (in SVHN).
    \item Number of epochs for warm start before the tuning begins: 20 (in MVTecAD) and 40 (in SVHN).
    \item Number of updates for detector parameters $\theta$ per each update of $\mathbf{a}$: 1 (in MVTecAD) and 5 (in SVHN).
    \item Number of maximum training iterations: 500 (in MVTecAD) and 100 (in SVHN).
\end{compactitem}

Batch size is small in MVTecAD, since the images in the dataset have high resolution $256 \times 256$, causing high memory cost, while the number of samples is small in both training and test data.
The number of epochs for warm start and the number of updates for $\theta$ are set large in SVHN, since it has more diverse images than in MVTecAD and requires more updates of $\theta$ to decrease the training loss $\trnLoss$ sufficiently.
The number of maximum training iterations is set differently so that the total number of updates for $\theta$ is the same in both datasets.
Please find more details from our code repository: \url{https://github.com/jaeminyoo/ST-SSAD}.

It is noteworthy that the choice of those hyperparameters is done by observing the training process, especially how fast the training loss $\trnLoss$ decreases before and during the iterations, rather than the actual performance of \method given a labeled dataset, which is not accessible in unsupervised anomaly detection tasks.

\subsection{Augmentation Data Sizes}

Our validation loss $\valLoss$ is designed to be robust to $|\augData| / |\trnData|$, since the ratio of true anomalies is unknown at training time.
Consequently, we simply set $|\augData| = |\trnData|$ in all experiments, such that we perform one augmentation per sample. 
At the same time, instead of using all training data in every computation of $\valLoss$, we randomly sample 256 training samples (and 256 augmentation samples) at each computation for efficiency. The number 256 is chosen large enough to estimate the true distribution of training data.

To verify the robustness of $\valLoss$, we varied the number of augmentation samples used in the computation of $\valLoss$ to 64, 128, 256, and 512, with 256 being the choice in the original experiments.
The performance remains stable across different augmentation sizes, yielding similar results on various tasks.
Specifically, on the eight anomaly types of the Cable object in MVTecAD, the four settings resulted in average ranks of $2.6$, $2.3$, $2.5$, and $2.4$, respectively.